\documentclass[11pt]{article}
\usepackage[preprint]{acl}
\usepackage{times}
\usepackage{latexsym}
\usepackage[T1]{fontenc}
\usepackage[utf8]{inputenc}
\usepackage{microtype}
\usepackage{inconsolata}
\usepackage{graphicx}
\usepackage{float}
\usepackage[table]{xcolor}
\usepackage{array,booktabs,amsmath,amssymb,multirow,url}

\title{ColorRef: Calibrating Verbal Feedback for Color Revision}
\author{Pablo Loyola\textsuperscript{1}, Andres Hoyos-Idrobo\textsuperscript{2}, Edison Marrese-Taylor\textsuperscript{3}, Colin Rowat\textsuperscript{1}, Alfredo Solano\textsuperscript{3} \\
\textsuperscript{1} Rakuten Group, Inc; Tokyo, Japan\\
\textsuperscript{2} Rakuten Group, Inc; Paris, France\\
\textsuperscript{3} Graduate School of Engineering, The University of Tokyo\\
\{pablo.a.loyola,andres.hoyosidrobo, colinian.rowat \}@rakuten.com, \{emarrese, asolano\}@weblab.t.u-tokyo.ac.jp
}

\begin{document}
\maketitle

\begin{abstract}
People rarely specify an exact color when asking someone to revise a design:
they might say \textit{a little darker} or \textit{more blue} after seeing
a proposal. A language-model assistant must then infer not only which way
to change the color, but how far. We study this distinction with
\textsc{ColorRef}, a color reference game in which a model guesses a color
from a description and a teacher, knowing a recorded target hidden from the
guesser, supplies verbal corrections. On 1{,}000 crowd-sourced
description--target pairs, three rounds of deterministic oracle feedback
reduce mean target distance from 42.33 to 22.15 $\Delta E_{76}$.
However, those corrections provide additional information, and improvement
alone does not establish well-calibrated feedback following. A controlled
Qwen3-14B diagnostic reveals a more specific failure: following a
requested direction and ordering graded phrases do not guarantee a
suitable step size; strong modifiers can induce excessive changes. We therefore measure
each receiver's typical response to \textit{a little}, \textit{somewhat},
and \textit{much} on separate colors and select wording matched to its
observed steps. On 128 held-out starting colors, calibration reduces
one-step mean error from 12.36 to 6.86 for Qwen3-14B and from 11.06
to 5.89 for Gemma 4 12B against a fixed graded-wording rule.
The benefit is concentrated on large corrections, with many unchanged
cases and no consistent increase in threshold success.
Our findings show that following the direction of feedback is not the
same as making an appropriately sized revision.
\end{abstract}

\begin{figure*}[t]
\centering
\begin{minipage}[t]{0.41\textwidth}
\vspace{0pt}
\centering
\textbf{(a) Reference-game exchange}\par\smallskip
\includegraphics[width=\linewidth]{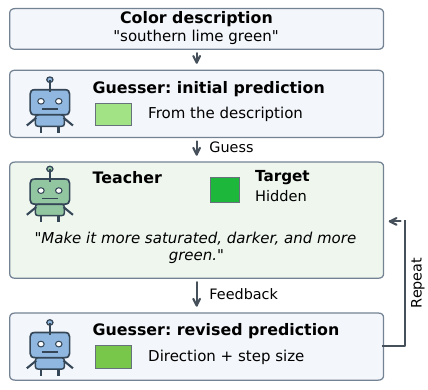}
\end{minipage}\hfill
\begin{minipage}[t]{0.57\textwidth}
\vspace{0pt}
\centering
\textbf{(b) Saved color trajectory}\par\smallskip
\includegraphics[width=\linewidth]{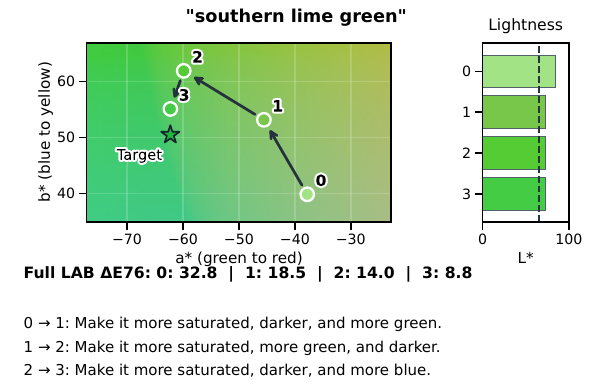}
\end{minipage}
\caption{ColorRef for \textit{southern lime green} (example 427046).
(a) The guesser predicts from the description; the teacher knows the
hidden target and gives feedback for revision.
(b) Qwen3-14B with an axis oracle: point 0 is the initial prediction,
points 1--3 are fixed-round revisions, and the star is the target.
The $a^*b^*$ plane shows chromatic movement; the strip shows $L^*$
(target dashed). Lower rows give full $\Delta E_{76}$ and exact feedback.
Both panels use the same saved game. This is an illustration, not an
aggregate result; plotting details appear in Appendix~\ref{app:state_protocol_details}.}
\label{fig:colorref_overview}
\label{fig:intro_game}
\label{fig:game_trajectory}
\end{figure*}

\section{Introduction}
\label{sec:introduction}

Color is a continuous perceptual domain that people describe through
language \cite{cie2018colorimetry,mcmahan2015bayesian}. Descriptions can
name properties, as in \textit{dark green}, refer through familiar objects,
as in \textit{mustard yellow}, or evoke associations, as in
\textit{abandoned hope} \cite{loyola-etal-2023-perceptual}.
They need not identify a unique color: interpretation can vary with
context, speaker judgments, and subjective associations
\cite{mcmahan2015bayesian,monroe2017colors}.

This makes color a useful setting for studying grounded language in
large language models \cite{abdou-etal-2021-language}.
A \emph{one-shot prediction} captures a model's initial interpretation of
a description, but communication can continue through clarification and
revision \cite{clarkbrennan1991grounding,benotti2021recipe}.
When a proposed color differs from what a speaker intends, a correction
such as \textit{a little darker} or \textit{more blue} can guide the next
guess \cite{winn-muresan-2018-lighter}. How effectively can a model use
such feedback to move toward the intended color?

We study this process through \textsc{ColorRef}, building on
speaker--listener color reference games \cite{monroe2017colors}
(Figure~\ref{fig:colorref_overview}). Each description is paired with
a crowd-sourced \emph{recorded target color}. The \emph{guesser} sees the
description, not a target swatch, and proposes a color as an sRGB HEX code
in the main game. The \emph{teacher} sees both the target and the current
guess and sends verbal feedback; the target itself remains hidden from
the guesser. The guesser then receives the description, its previous
displayed guess, and the latest feedback, and produces a revised color.
A deterministic \emph{oracle teacher} selects correction directions
directly from the known target--guess discrepancy; the oracle's privileged
information is the known target, not an optimal teaching policy.
Exchanges continue for a fixed revision budget or a specified
stopping rule (Section~\ref{sec:game}).

For example, the source corpus records \textit{faded old rose} with target
\texttt{\#D2917A}; the guesser must propose a color from the description
alone before hearing corrections. The recorded pairing is an evaluation
referent, not the only semantically acceptable interpretation of that name.
The sequence of guesses makes receiving corrections observable separately
from initial description interpretation and teacher message production.
We quantify closeness to a recorded target using $\Delta E_{76}$:
Euclidean distance between the target and displayed guess in CIELAB
coordinates, with smaller values denoting closer agreement. This is
not a measurement of whether a color name has a unique human meaning.

A correction has two aspects. \emph{Direction} specifies which property
should increase or decrease, as in \textit{lighter} or \textit{more blue}
\cite{winn-muresan-2018-lighter}. \emph{Magnitude} specifies the extent:
\textit{a little darker} and \textit{much darker} express different degrees
of adjustment \cite{kennedymcnally2005scale}.
Recent work probes LLM understanding of scalar modifiers
\cite{gao-etal-2025-masp} and their translation of vague intensity words
into numeric actions \cite{tabach2026slightly}.
Color revision poses a concrete version of this question: how much should
the model change a color when asked to make it \textit{a little darker}?
Knowing that \textit{a little} requests less change than \textit{much}
does not specify a useful step size; an ordered, directionally correct
update can still undershoot or overshoot the target
(Section~\ref{sec:metrics}). We test direction following, graded step
sizes, and whether model-produced feedback preserves supplied corrections
and helps revision.

Additional information should generally help a target-aware teacher
correct a guess; the more discriminating question is whether following a
linguistically appropriate direction yields a useful \emph{amount} of
movement. We therefore combine description-based games with supplied-start
comparisons and isolated color-adjustment tasks. These controls distinguish
initial interpretation, directional compliance, and step-size accuracy,
without claiming to recover the models' internal perceptual representations.

Our main intervention is \emph{receiver-calibrated feedback}: on separate
calibration colors, we measure the typical displayed step induced by
each graded phrase and direction. A target-aware teacher selects the
phrase whose measured step is closest to the remaining correction. The
policy uses the same three graded expressions as a fixed, unfitted wording
rule, but adapts phrase selection to the receiving model's observed
response. We independently calibrate Qwen3-14B and Gemma 4 12B and
evaluate them on the same 128 held-out starting colors
(Section~\ref{sec:setup_magnitude}).

\paragraph{Main findings.}
Our experiments support four observations:
\begin{enumerate}
\item \textbf{Target-aware feedback helps, but does not solve grounding.}
On 1{,}000 balanced description--target pairs, three oracle revisions
reduce mean recorded-target error from 42.33 to 22.15 $\Delta E_{76}$;
only 38 final guesses reach the operational error threshold of five.
These revisions add target information, so this is not an
information-matched comparison of one-shot and multi-turn reasoning.
\item \textbf{Correct direction does not imply useful magnitude.}
In controlled color adjustments, models often follow the requested
axis and order graded words correctly, yet a strong modifier can
produce an excessive step and increase target error.
\item \textbf{Measured receiver behavior improves phrase selection.}
Independently calibrated Qwen3-14B and Gemma 4 12B policies reduce
held-out one-step mean error against the same fixed graded-wording rule.
Improvements concentrate on large corrections; many paired responses
are identical, and threshold convergence need not increase.
\item \textbf{Teacher fidelity and receiver benefit differ.}
Oracle-assisted LLM teachers can preserve supplied direction sets,
especially with demonstrations, without necessarily improving the
receiver's target accuracy over canonical oracle wording.
\end{enumerate}
Our contribution is to show why two common checks of instruction
following---moving in the requested direction and increasing steps
for stronger modifiers---need not predict a useful correction.
ColorRef measures this gap in color revisions and evaluates a
receiver-specific wording policy against a fixed rule. This is a
controlled result for single-axis adjustments, not evidence that
the gap generalizes to other agents or that calibration improves
full description-based games.

\section{Related Work}
\label{sec:related_work}

\paragraph{Color meaning and perceptual representations.}
Color is a well-studied meeting point between continuous perception
and discrete language. Work on basic color terms and cross-linguistic
naming patterns asks how languages partition color space
\citep{berlin1991basic,regier2007color}; communicative-efficiency
accounts additionally examine the tradeoff between distinctions conveyed
and the complexity of a naming system \citep{zaslavsky2018efficient}.
These questions concern regularities of color language across speakers
and languages, not whether a particular name corresponds to a uniquely
correct point. Indeed, a description can identify a broad region or
depend on context. Grounded semantic approaches model such variation
in referring expressions and perceptual observations
\citep{mcmahan2015bayesian,monroe2017colors}.

A related line of work asks whether language-model representations reflect
perceptual color geometry even without direct visual grounding.
\citet{abdou-etal-2021-language} compare language-model color-term
representations with perceptual structure, while
\citet{loyola-etal-2023-perceptual} examine how these relationships
change with abstract and subjective color descriptions.
Such representational comparisons motivate color as a diagnostic domain,
but they do not by themselves determine how a model will \emph{act}
when asked to revise a particular displayed color. We study observable
revisions instead of inferring the existence or topology of an internal
perceptual space from task performance.

\paragraph{Reference games, comparatives, and interactive grounding.}
Pragmatic accounts treat meaning in a reference game as arising partly
from what a speaker expects a listener to infer
\citep{lewis1969convention,frank2012predicting,goodman2016pragmatic}.
\citet{monroe2017colors} develop a speaker--listener model for interpreting
color descriptions in context, demonstrating the value of pragmatic
reasoning for ambiguous reference. Comparatives provide a complementary
form of communication: rather than naming a new target directly, a
speaker can specify a change relative to a known state.
\citet{winn-muresan-2018-lighter}, for example, ground comparative
color descriptions as directions from a reference color.
ColorRef extends this operational perspective from whether the
direction is appropriate to whether the \emph{distance moved} is useful.
Its teacher knows an assigned target, and its guesser repeatedly
updates a displayed color; it is not a test of human dialogue or of
whether independent LLM agents establish shared meaning.

Clarification and grounding research emphasizes that communication
can continue after an initially incomplete interpretation
\citep{clarkbrennan1991grounding,benotti2021recipe}.
Recent interactive evaluation also distinguishes task success from
other dimensions of common-ground formation
\citep{imai2025measuring}. These distinctions matter here because
arriving closer to a target, obeying an instruction, and producing
a faithful correction message are different behavioral outcomes.

\paragraph{Graded language as quantitative instruction.}
A phrase such as \textit{somewhat darker} expresses a degree, but
degree-modification semantics does not supply a universal numeric
step in CIELAB space \citep{kennedymcnally2005scale}.
Recent LLM studies probe the interpretation of scalar modifiers
\citep{gao-etal-2025-masp} and examine vague intensity language
translated into numeric actions \citep{tabach2026slightly}.
We ask a narrower intervention question: given a fixed color-adjustment
prompt and a vocabulary of three graded expressions, can a teacher
choose a phrase using a receiver's \emph{measured} movement for that
direction? We evaluate this selection rule against a specified
unfitted rule on held-out starting colors. The approach is a
receiver-specific operational calibration, not a general semantic
definition of these words.

\paragraph{Feedback, critiques, and iterative revision.}
Human and AI feedback are widely used to train and improve language models
\citep{ouyang2022training,bai2022constitutional,saunders2022selfcritiquing}.
At inference time, methods such as Self-Refine and Reflexion use verbal
critique or reflection to improve subsequent outputs
\citep{madaan2023selfrefine,shinn2023reflexion}. Their broader settings
make it difficult to isolate which information in a message causes an
improvement. ColorRef deliberately restricts the output to a measurable
color state and, in its controls, supplies exact starting states and
correction directions. This lets us distinguish the fidelity of the
message a teacher generates from the accuracy of the receiver's
subsequent action. Our teacher experiments concern restatement of
oracle-supplied directions, not autonomous critique quality.

\section{Framework}
\label{sec:colorref}

ColorRef consists of a color-description task and a sequence of
feedback-driven revisions. The description establishes an initial guess;
the teacher's corrections identify how the current color differs from
a recorded target. We then use controlled tasks, in which starting
colors and correction directions are supplied, to examine the response
to feedback without confounding it with interpretation of the name
(Figure~\ref{fig:colorref_overview}).

\subsection{Descriptions and assigned targets}
\label{sec:dataset}
\begin{table}[t]
\centering\small
\setlength{\tabcolsep}{3pt}
\renewcommand{\arraystretch}{1.05}
\newcommand{\exampleswatch}[1]{\textcolor[HTML]{#1}{\rule{1.8em}{.9em}}}
\begin{tabular}{@{}llc@{}}
\toprule
Regime & Description & Recorded target \\
\midrule
\multirow{2}{*}{\shortstack[l]{Explicit\\grounded}}
 & dark lavender shoot & \exampleswatch{A44A8B} \\
 & white sky & \exampleswatch{A8DAEC} \\
\midrule
\multirow{2}{*}{\shortstack[l]{Prototype\\mediated}}
 & druidic moss & \exampleswatch{378421} \\
 & light stop sign & \exampleswatch{FD431C} \\
\midrule
\multirow{2}{*}{\shortstack[l]{Compound\\associative}}
 & newest green bud & \exampleswatch{91E45C} \\
 & orangutan reddish & \exampleswatch{F5240A} \\
\midrule
\multirow{2}{*}{\shortstack[l]{Abstract\\idiosyncratic}}
 & dreamy kiss & \exampleswatch{F374A5} \\
 & celestial mistress & \exampleswatch{95082E} \\
\bottomrule
\end{tabular}
\caption{Two illustrative descriptions per parser-based regime; swatches show their
recorded sRGB targets. These pairings do not exclude other plausible
interpretations.}
\label{tab:regime_examples}
\end{table}

Our descriptions and recorded color targets come from
colornames.org\footnote{\url{https://colornames.org/}}, a collection
of user-provided names associated with sRGB HEX colors, not a
standardized color dictionary such as Pantone. A retained example
$(x_i,h_i,c_i,r_i)$ contains a description $x_i$, its recorded
HEX target $h_i$, the corresponding D65 CIELAB coordinate $c_i$,
and a coarse parser-assigned semantic regime $r_i$.
CIELAB represents lightness by $L^*$, green--red variation by
$a^*$, and blue--yellow variation by $b^*$.

The retained English corpus has 518{,}830 distinct descriptions.
A GPT-4o-based description-validity score filters out strings below
0.75, retaining plausible color names rather than guaranteeing agreement
about their recorded targets. This filter is not human validation of
color-name meaning. Table~\ref{tab:regime_examples} illustrates the four categories.
\emph{Explicit grounded} names have recognizable color anchors,
such as \textit{dark lavender shoot}.
\emph{Prototype mediated} names invoke objects or materials, such
as \textit{druidic moss}. \emph{Compound associative} names combine
color cues and other references, as in \textit{newest green bud}.
\emph{Abstract idiosyncratic} names, such as \textit{dreamy kiss},
rely more on subjective or figurative associations.

The categories are assigned by a deterministic lexical parser,
which searches for basic color words, modifiers, and prototype terms
and uses heuristic scores for less direct descriptions. They enable
stratified analysis but should not be read as human-validated semantic
classes or as a continuous measurement of directness. In particular,
a larger error against a single assigned target is not proof that a
description has been misunderstood. We evaluate agreement with the
recorded pair rather than a uniquely correct human interpretation.
The complete preprocessing and parser details are retained in
Appendix~\ref{sec:appendix-dataset}.

\subsection{Interactive states and evaluation}
\label{sec:game}
At turn zero, the guesser receives only the description $x_i$ and
outputs a color. This is the paired description-only baseline.
For each subsequent turn, the teacher sees the current displayed
guess and recorded target and chooses verbal corrections. The
guesser sees the description, its previous displayed color, and the
latest feedback, but never the target or a rendered target swatch.
Prompts use a fresh context at each revision rather than preserving
the entire preceding dialogue.

Consider a guess that is too light and too far toward red.
The deterministic teacher may say \textit{Make it darker and more green}.
The guesser must then output another color, which may or may not
move in both requested directions. In the main description-based games,
an oracle selects up to three qualifying correction directions from
lightness, red--green, yellow--blue, and HSV saturation discrepancies.
These are qualitative requests, not numeric constraints on the output
and not graded modifiers such as \textit{much}. The three-revision
evaluation continues for its fixed budget, even if a guess becomes
close to the target. Hence the first guess and subsequent revisions
can be compared, but the additional oracle information must not be
mistaken for the effect of interaction alone.

To evaluate color changes consistently, we distinguish the
model's \emph{native output} from the \emph{displayed color}.
HEX predictions already specify a finite-resolution sRGB color.
For a native LAB output $z_t$, we clip and round its conversion to
8-bit sRGB, then convert that displayed color back to CIELAB,
giving $\hat c_t=P(z_t)$. The oracle, the next prompt, and the
primary target-error metric use $\hat c_t$; the original LAB output
is retained to diagnose out-of-gamut projection or coordinate errors.
Thus a seemingly valid numeric change can be altered when rendered
as a displayable color. Additional state and figure conventions
appear in Appendix~\ref{app:state_protocol_details}.

\subsection{Receiver-calibrated wording}
\label{sec:calibration}
The description-based oracle asks for directions. Our central
intervention additionally considers \emph{how much} the receiver
moves when asked for a graded adjustment. We first isolate this
question from color-name interpretation: for a supplied starting
CIELAB color, the receiver gets a single-axis instruction such as
\textit{make it somewhat more red} and returns a new LAB color.
The tested directions are lighter, darker, redder, greener,
yellower, and bluer; each is paired with \textit{a little},
\textit{somewhat}, or \textit{much}.

We measure typical responses on separate calibration colors.
For a starting displayed color $b_j$, a parsed displayed
response $\hat c'_{j,d,q}$, a signed unit direction $d$, and
a phrase $q$, the median requested-axis movement is
\begin{equation}
m_{d,q}=\operatorname{median}_{j\in J_{d,q}}
 d^\top(\hat c'_{j,d,q}-b_j).
\end{equation}
Here $d$ makes movement in the requested direction positive.
The median includes parsed zero and wrong-direction movements;
unparsed responses are counted separately rather than silently
repaired. The resulting table estimates what \emph{this receiver
under this prompt} tends to do, not what a phrase means to humans.

When the teacher knows a target $c$ and current displayed color
$\hat c$, it can compute the remaining change along a chosen
direction, $\rho=d^\top(c-\hat c)$. It selects
\begin{equation}
q^*=\arg\min_{q\in Q_d}|m_{d,q}-\rho|,
\end{equation}
where $Q_d$ contains the three graded phrases with usable positive
medians. The selected wording, not the numeric residual or
calibration table, is sent to the receiver.

\paragraph{A concrete example.}
For Qwen3-14B, the confirmation-study median \emph{redward}
steps are approximately 1.1, 6.8, and 76.3 LAB units for
\textit{a little}, \textit{somewhat}, and \textit{much},
respectively. Suppose a target lies six units farther in
the red direction. The fixed comparator, with cutpoints
at 9 and 18, requests \textit{a little more red};
the measured-response policy instead requests
\textit{somewhat more red}, because its typical 6.8-unit
step is closer to the desired six than a 1.1-unit step.

Now suppose the requested redward correction is 24 units.
The fixed rule selects \textit{much more red}, although the
observed median step for that phrase is about 76.3 units.
The calibrated policy selects \textit{somewhat more red}:
the differences from the desired 24-unit change are
17.2 versus 52.3 units, respectively. This does not make
\textit{somewhat} an exact 24-unit instruction. It avoids
a predicted large overshoot at the cost of leaving a
residual correction, illustrating the limited resolution
of the three-phrase vocabulary.

These are illustrative calculations from calibration medians,
not the measured endpoints of any individual held-out case.
Actual responses can also move along unintended axes or change
after sRGB projection. We freeze each model's calibration mapping
before held-out evaluation; we do not fine-tune its weights.
The exact-numeric comparison supplies a specified coordinate
change and holds other coordinates fixed, so it is an execution
control with more information than verbal wording. Calibration
tie handling and missing mappings are detailed in
Appendix~\ref{app:calibration_selection_details}.

\subsection{Metrics and interpretation}
\label{sec:metrics}
We quantify agreement with the recorded target by
$e_t=\|c-\hat c_t\|_2$, the Euclidean CIELAB distance
$\Delta E_{76}$ between target and displayed prediction.
Lower values mean closer agreement with that one recorded
color, not necessarily a more acceptable interpretation to a
human speaker. The \emph{gain} over a game is $e_0-e_T$;
positive gain indicates an improvement from the starting
guess. We call $e_T\leq5$ \emph{convergence} for evaluation,
but this is a fixed operational threshold rather than a
universal perceptual visibility criterion.

Target error differs from both the size and direction of an
update. Let $v=c-\hat c_t$ be the correction still needed
and $u=\hat c_{t+1}-\hat c_t$ the observed displayed
movement. The response may follow a requested axis
(\emph{sign compliance}) or point generally toward the
target (\emph{alignment}) without getting closer. Indeed,
\begin{equation}
\|v-u\|_2^2-\|v\|_2^2
=\|u\|_2^2-2v^\top u.
\end{equation}
Even when $v^\top u>0$, the new error increases if the
step is sufficiently large. This is why a model can
understand \textit{more red} directionally and still
produce a harmful revision.

For the controlled adjustment studies we additionally
measure requested-axis step size, movement on other axes,
whether \textit{a little}, \textit{somewhat}, and
\textit{much} elicit ordered responses, and disagreement
between native and displayed LAB states. These diagnostics
distinguish ordinal language following from useful
target-directed correction. We use $\Delta E_{76}$ for
reproducibility and its approximate perceptual geometry,
without assuming perfect perceptual uniformity; further
definitions appear in Appendix~\ref{app:metric_definitions}.

\section{Experimental Setup}
\label{sec:experimental_setup}

Our evaluation separates four questions that would otherwise be
mixed together. First, do target-aware corrections reduce the error
of color guesses made from descriptions? Second, does correction
behavior depend on whether the model outputs HEX or CIELAB?
Third, when starting colors and targets are supplied directly,
does selecting graded wording from measured receiver responses improve
one-step correction over a fixed wording rule? Finally, can a
language-model teacher preserve supplied correction directions,
and do those restatements help the receiver? The third question
is our primary policy comparison; the others establish context
and expose competing explanations. In particular, we do not test
a fully calibrated teacher in the description-based reference game.

Qwen3-14B \citep{yang2025qwen3} is used throughout, with Gemma 4 12B
\citep{gemmateam2026gemma4} independently replicating the one-step
magnitude comparison. Both use Hugging Face Transformers
\citep{wolf2020transformers}, temperature zero, thinking disabled,
and fresh contexts. Output budgets are 64 tokens for receivers and
80 for teachers. Inputs, prompts, and model revisions are frozen
before evaluation; detailed model revisions and complete task
templates are included in
Appendices~\ref{app:execution_settings} and~\ref{app:all_prompts}.

\subsection{Does feedback help description-based guesses?}
The corpus is a sampling pool, not a set of 518{,}830 uniquely verified
semantic targets: evaluating every entry would not resolve target
ambiguity. We freeze 1{,}000 descriptions, stratified to 250 per regime,
to examine the four parser categories at equal weight. Consequently,
pooled results describe this balanced evaluation, not a corpus-frequency
estimate. This sample also cannot measure inter-speaker target agreement.
Games use HEX output and three fixed revisions, with each initial prediction
as its paired description-only baseline. Magnitude tests instead use
separate held-out starting colors and supplied coordinate targets.
The deterministic oracle selects up to three corrections from LAB
lightness, red--green, yellow--blue, and HSV saturation discrepancies,
using fixed thresholds and ranking. A selected direction is a
\emph{constraint}; the maximum count is feedback \emph{bandwidth},
not step size. Canonical feedback, such as \textit{Make it darker and
more green}, contains no graded magnitude modifiers.
Appendix~\ref{app:grounding_protocol} specifies sampling and oracle selection.

\subsection{Does output encoding change correction?}
\label{sec:setup_interface}
On 64 descriptions, 16 per regime, we compare HEX, plain LAB, and LAB
with an explicit axis legend. All three receive the same displayed start,
drawn from earlier HEX predictions, and the same first oracle correction.
The first revision is the primary matched endpoint; the third fixed
revision is secondary. Later feedback adapts to each arm's state and can
differ. This separates correction from each interface's initial color
prediction. Display conversion follows Section~\ref{sec:game}; detailed
protocols appear in Appendix~\ref{app:new-controls}.

\subsection{Does calibration improve magnitude selection?}
\label{sec:setup_magnitude}
Isolated probes first compare six directions and four wordings
(bare, \textit{a little}, \textit{somewhat}, \textit{much}) on 12 fixed
colors under plain and axis-legend LAB prompts. Exact 5-/10-unit and
no-change controls distinguish coordinate execution from qualitative
adjustment (Appendix~\ref{app:new-controls}).

The main confirmation uses axis-legend LAB prompts, fits the mapping in
Section~\ref{sec:calibration} on 32 calibration colors, and evaluates
128 disjoint starting colors.
Targets are constructed by moving one native LAB coordinate by 6, 12,
or 24 units; projecting them into displayable sRGB can change the
corresponding displayed residual. Frozen input-only feasibility checks
retain 2{,}108 targets;
generated responses are never filtered by projection error. Each case
has four arms from the same start, each producing one LAB revision:
\emph{bare} direction; \emph{unfitted} graded wording using fixed residual
cutpoints 9 and 18; \emph{calibrated} wording using fitted receiver
medians; and an \emph{exact numeric} instruction with coordinate holds.
The primary contrast is calibrated minus unfitted displayed target error.
Numeric feedback is an execution control with greater precision.

Gemma repeats the identical frozen input plan, task order, and prompts,
fitting its own mapping on the same 32 calibration colors. Previously
used colors are excluded before inference; neither model's held-out
responses fit its policy. The 128 starting colors are the sampling units;
their multiple direction/distance targets are repeated observations.

\subsection{Transfer and teacher-message diagnostics}
\label{sec:setup_teacher}
An exploratory Qwen pilot fits medians on 12 colors, then tests sequential
transfer on 12 fresh starts and 200 feasible targets. Bare, calibrated,
and numeric arms stop at target-known displayed $\Delta E_{76}\leq5$,
with a five-revision cap. The original axis stays fixed, but the requested
sign can reverse after overshoot (Appendix~\ref{app:new-controls}).

Teacher studies use 60 shared-start examples, comparing natural and
restricted restatement prompts, each zero-shot or four-shot, at bandwidths
one and three. All receive the same states and supplied oracle directions;
Qwen3-14B serves as both the language-model teacher and the HEX
receiver in this experiment. Every resulting message is passed unchanged
to the receiver alongside a canonical-oracle arm: 120 cases with
five feedback conditions.
Restatement audits and receiving accuracy are assessed separately;
no audit selects or repairs messages. These are oracle-assisted
restatement tests, not unaided teacher geometry
(Appendix~\ref{app:teacher_controls}).

\paragraph{Uncertainty and failures.}
Paired percentile bootstrap intervals resample whole descriptions or
examples within regimes, and whole starting colors for magnitude studies,
keeping associated cases together. Intervals condition on observed cohorts, not generation randomness.
Strict parse failures remain failures, without repaired colors; backend
exceptions remain pending. Accuracy uses explicitly matched parsed
cohorts with available-pair sensitivity. Completion and failure counts
accompany results. Subgroup analyses are exploratory.

\section{Results}
\label{sec:results}
Description-based games establish whether feedback helps; shared-start
and isolated controls examine direction and magnitude. We then test
receiver calibration and distinguish teacher restatement from receiving
accuracy. Tables report displayed-state target error throughout.

\subsection{Verbal correction narrows target disagreement}
\label{sec:results_main}
All 4{,}000 responses parse, yielding paired outcomes for all 1{,}000
descriptions. Mean error falls from 42.33 [40.70, 44.01] to 22.15
[21.28, 22.99], with paired gain 20.19 [18.61, 21.79] $\Delta E_{76}$.
This is a 47.7\% reduction in mean error, not mean per-example relative
improvement. Convergence rises from 5 to 38 of 1{,}000; most final
guesses remain outside the threshold.

\begin{table}[t]
\centering\small
\begin{tabular}{@{}lrrr@{}}
\toprule
Regime & Initial $\Delta E$ & Final $\Delta E$ & Gain \\
\midrule
\shortstack[l]{Explicit\\grounded} & 36.41 & 22.02 & 14.39 \\
\shortstack[l]{Prototype\\mediated} & 43.39 & 22.41 & 20.98 \\
\shortstack[l]{Compound\\associative} & 37.89 & 20.64 & 17.24 \\
\shortstack[l]{Abstract\\idiosyncratic} & 51.64 & 23.51 & 28.12 \\
Overall & 42.33 & 22.15 & 20.19 \\
\bottomrule
\end{tabular}
\caption{Fresh Qwen3-14B grounding evaluation: 250 descriptions per regime,
canonical $c3$ feedback and three fixed revisions. Initial predictions are
the paired one-shot baseline. Gain is initial minus final error.}
\label{tab:main_results}
\end{table}

All regimes improve descriptively. Abstract descriptions start with the
greatest disagreement and show the largest absolute gain
(Table~\ref{tab:main_results}). The gain should not be interpreted as
stronger feedback-following for abstract language: those examples begin
farther from their recorded targets and have more room to improve.
The reported means do not control for initial error or establish a
between-regime effect. Their final mean is closer to those of the other
regimes, suggesting that an initially difficult name need not preclude
subsequent localization of an assigned target. Nevertheless, only
38 of 1{,}000 final predictions reach the operational
$\Delta E_{76}\leq5$ threshold. These results demonstrate
target-directed correction under additional oracle information, not
that the guesser has inferred a unique human meaning or that multiple
exchanges outperform an information-matched one-shot instruction.

\subsection{Output encoding affects correction from shared starts}
\label{sec:results_interface}
The shared start separates receiving behavior from how each interface
initially interprets the description: the same displayed color and first
correction are supplied in all arms.
All 576 revisions parse across the 64 shared-start examples.
Plain LAB exceeds HEX first-revision error by 6.19 [0.85, 11.89]
$\Delta E_{76}$ (Table~\ref{tab:shared_interface}). The axis legend
reduces plain-LAB error by 5.71, although its first-revision interval
spans zero. After three revisions, legend minus plain LAB is $-12.11$
[$-18.04$, $-5.49$]. Legend minus HEX is 0.38 [$-4.85$, 6.29], which
does not establish equivalence. First corrections match; later feedback
is adaptive, and the comparison is conditional on HEX-derived starts.
Thus, requesting coordinates does not by itself make correction easier;
the instructions accompanying that interface matter as well.

\begin{table}[t]
\centering\small
\begin{tabular}{@{}lrrr@{}}
\toprule
Interface & First $\Delta E$ & Third $\Delta E$ & C-sat \\
\midrule
HEX & 32.10 & 22.78 & 0.888 \\
Plain LAB & 38.29 & 35.28 & 0.823 \\
LAB + legend & 32.58 & 23.16 & 0.906 \\
\bottomrule
\end{tabular}
\caption{Shared starts and first corrections, $N=64$ per interface;
mean starting error 41.96. C-sat is first-revision constraint satisfaction.
Displayed-state error is primary. Three revisions
are forced rather than stopped at convergence.}
\label{tab:shared_interface}
\end{table}

\begin{figure*}[t]
\centering
\includegraphics[width=0.98\textwidth]{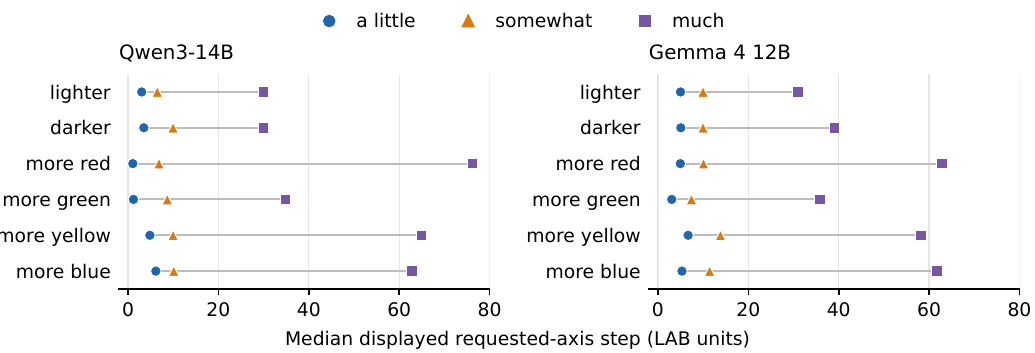}
\caption{Fitted phrase responses for the larger confirmation and replication:
median displayed requested-axis steps on the same 32 calibration colors,
measured separately for each model. Each point uses 32 parsed responses,
except Qwen's \textit{a little more red} (31). The shared linear scale shows
ordered but unevenly spaced steps across directions and receivers.
These are calibration medians, not held-out target errors, uncertainty
intervals, or human perceptual units.}
\label{fig:calibration_steps}
\end{figure*}

\subsection{Direction and graded magnitude are separate requirements}
\label{sec:results_quantifiers}
All 888 isolated-probe responses parse. Axis instructions increase
positive-direction qualitative updates from 228/288 to 288/288,
strictly increasing comparisons from 172/216 to 210/216, and
nondecreasing comparisons from 208/216 to 216/216. These overlapping
comparisons involve 12 fixed anchors, not 216 independent colors.
Numeric and no-change controls are exact except one legend-condition
10-unit instruction executed as a 20-unit update.

Despite this ordering, pilot displayed medians vary across directions:
1.14--4.92 for \textit{a little}, 5.07--10.09 for \textit{somewhat},
and 28.72--75.84 for \textit{much}. Of 72 \textit{much} calibration
responses, 54 have projection error above one. Direction following and
ordinal ordering therefore do not ensure a suitable displayed step;
these measurements motivate receiver-specific phrase selection.
Ordering asks whether a stronger modifier elicits a larger step.
Target accuracy instead asks whether that step fits the remaining
correction. A response can satisfy the first requirement while moving
past the target.

\subsection{Receiver calibration improves held-out phrase selection}
\label{sec:results_magnitude}
The larger calibration splits show why the choice of modifier matters
(Figure~\ref{fig:calibration_steps}). Both receivers' median ladders are
ordered, but the gaps depend on direction. For Qwen, \textit{much} is
about a 30-unit displayed step for lightness and a 76-unit step toward
red; Gemma's corresponding red median is about 63. These measurements
fit the selection policy before held-out evaluation, so the test asks
whether adapting wording to the receiver improves subsequent correction.

\begin{table*}[t]
\centering\small
\setlength{\tabcolsep}{5pt}
\begin{tabular}{@{}lrrrrrl@{}}
\toprule
Model & $N$ & Bare & Unfitted & Calibrated & Numeric &
\shortstack{Calibrated $-$ unfitted\\$\Delta$ error [95\% interval]} \\
\midrule
Qwen3-14B & 2{,}098 & 11.91 & 12.36 & 6.86 & 0.35 &
$-5.50$ [$-5.79$, $-5.20$] \\
Gemma 4 12B & 2{,}081 & 8.36 & 11.06 & 5.89 & 0.08 &
$-5.16$ [$-5.48$, $-4.84$] \\
\bottomrule
\end{tabular}
\caption{Mean displayed target error ($\Delta E_{76}$) and primary paired
contrast; negative differences favor calibration. $N$ counts common parsed
quartets, each model covering the same 128 starting colors with independently
fitted medians. Intervals resample whole starting colors. Numeric instructions
provide additional precision. Completion, costs, and exploratory distance
contrasts appear in Appendix~\ref{app:confirmation_display}.}
\label{tab:magnitude}
\end{table*}

Calibration lowers held-out error relative to the same fixed unfitted
rule for both models (Table~\ref{tab:magnitude}). Each completes all
8{,}432 responses. Qwen's bare/unfitted/calibrated/numeric parse failures
are 3/5/0/2; Gemma's are 0/0/0/27. Common parsed quartets cover all 128
starts, with model-specific denominators. Available language-pair
contrasts remain similar: $-5.53$ for Qwen ($N=2{,}103$) and $-5.10$
[$-5.42$, $-4.77$] for Gemma ($N=2{,}108$), without requiring all four
arms to parse. These are within-model effects, not a direct
between-model ranking.

Graded wording alone does not establish an improvement over bare
directions: Qwen's unfitted-minus-bare difference is 0.45
[$-0.69$, 1.52], while Gemma's is 2.69. Qwen's graded arms
follow the requested direction on roughly 99\% of parsed
responses, yet the fixed rule can yield worse mean target error
than the bare instruction. This is the central behavioral distinction:
a response can be linguistically ordered and sign-correct while
operationally too large. The numeric arm nearly reaches the
constructed targets when its response parses, but also specifies
exact coordinate changes and holds; it is an execution control,
not an information-matched natural-language baseline.

Exploratory distance effects concentrate the benefit at 24-unit targets:
calibrated-minus-unfitted error is $-18.35$ for Qwen and $-18.18$ for
Gemma. Harmful updates fall from 318 to 58 and from 333 to 75,
respectively. Yet the median paired difference is zero for both models:
many cases use identical wording and produce identical colors. Gemma's
convergence counts also remain nearly unchanged (1{,}400 versus 1{,}398).
For 24-unit red, yellow, and blue corrections, both fitted policies
replace \textit{much} with \textit{somewhat}; Gemma also does so for
darker corrections. Other directions retain \textit{much}, so calibration
does not simply weaken every large request.
Calibration chiefly avoids excessive large corrections rather than
improving every adjustment or guaranteeing threshold success.
The comparison establishes an advantage over the specified fixed
cutpoints, not over arbitrary hand-tuned policies or feedback
chosen by an optimal numerical controller. Distance, phrase-transition,
and identity audits appear in Appendix~\ref{app:confirmation_display}.

\subsection{A smaller pilot transfers calibration across revisions}
\label{sec:results_transfer}

\begin{table}[t]
\centering\small
\begin{tabular}{@{}lrrr@{}}
\toprule
Arm & Final $\Delta E$ & Converged & Calls \\
\midrule
Bare & 6.46 & 148 & 516 \\
Calibrated & 2.98 & 192 & 338 \\
Numeric & 0.10 & 200 & 205 \\
\bottomrule
\end{tabular}
\caption{Sequential Qwen pilot: 200 targets on 12 fresh starts,
target-known stopping at error $\leq5$, with a five-revision cap.
Calls exclude 288 calibration responses. Numeric instructions supply
additional precision; no unfitted graded arm is tested here.}
\label{tab:transfer_main}
\end{table}

The sequential Qwen pilot completes all 600 games with 1{,}059 parsed
revisions. On 200 targets from 12 fresh starts, calibrated minus bare
endpoint error is $-3.48$ [$-4.20$, $-2.74$]; convergence rises from
148 to 192, and evaluation calls fall from 516 to 338 (34.5\%, excluding
calibration; Table~\ref{tab:transfer_main}). All twelve per-start mean
contrasts favor calibration, although individual targets need not improve.
This target-known stopping pilot has no unfitted graded arm; the larger
confirmation tests that contrast for one revision only.

Eight calibrated trajectories exhaust the cap, all at 24-unit lighter
or greener targets with off-axis disagreement above threshold.
The fixed-axis policy does not explicitly correct display-induced drift.
Numeric feedback has the lowest endpoint error but includes five sequential
execution misses above one (maximum 44); subsequent oracle corrections,
not autonomous self-repair, bring those trajectories to convergence.
Full execution tails appear in Appendix~\ref{app:magnitude_diagnostics}.
The pilot therefore tests repeated use of a frozen response mapping,
while also showing where single-axis control fails: satisfying the
requested correction can leave disagreement on coordinates it does not
intentionally adjust.

\subsection{Restatement fidelity and usefulness are distinct}
\label{sec:results_llm_teacher}
All 480 teacher messages and 600 receiver revisions complete.
A supplementary scope-aware audit resolves all 120 direction sets in
each four-shot condition as preserved, versus 104/120 in restricted
zero-shot. Natural zero-shot has 81 matches, 17 mismatches, and 22
unresolved cases (Table~\ref{tab:teacher_receiver}). These are
finite-grammar diagnostics, not independent human judgments:
\textit{your guess is darker than the target} implies a lighter update,
while a harmless \textit{Feedback:} prefix can lower format certification.

\begin{table}[t]
\centering\small
\begin{tabular}{@{}lrrr@{}}
\toprule
Teacher arm & Matches & $\Delta E$ & C-sat \\
\midrule
Natural, 0-shot & 81 & 36.89 & 0.650 \\
Natural, 4-shot & 120 & 34.40 & 0.863 \\
Restricted, 0-shot & 104 & 31.94 & 0.867 \\
Restricted, 4-shot & 120 & 33.68 & 0.868 \\
Oracle & 120 & 33.78 & 0.868 \\
\bottomrule
\end{tabular}
\caption{Oracle-assisted direction restatement and one-revision reception,
120 cases on 60 examples. Matches are supplementary resolved set matches;
natural zero-shot has 22 unresolved cases. All receiver endpoints remain in
the comparison, with common starting error 39.83. C-sat uses supplied oracle
directions, not inferred teacher intent.}
\label{tab:teacher_receiver}
\end{table}

With original teacher text passed unchanged, restricted zero-shot lowers
receiver error relative to natural zero-shot by 4.95 [1.32, 8.39].
Natural four-shot's mean improvement is 2.49, with an interval spanning
zero. Restricted four-shot minus oracle is $-0.10$ [$-0.40$, 0.09],
without establishing equivalence.

Restricted zero-shot has the lowest mean but only 11 wins, 104 ties,
and five losses versus the oracle, with median difference zero.
Its advantage includes sixteen messages with added directions and is
concentrated in a few changed messages. Thus receiving accuracy is not a
fidelity ranking or an information-matched phrasing comparison.
The full decomposition appears in Appendix~\ref{app:teacher_controls}.

\section{Discussion}
\label{sec:discussion}
\paragraph{A game of interpretation and revision.}
ColorRef makes initial interpretation, feedback following, and feedback
production separately observable. The teacher fixes an intended referent
for an exchange; successful localization does not make that recorded
color the description's uniquely appropriate interpretation. Because
feedback adds information, the grounding gains support correction toward
an assigned target, not sequential adaptation alone. This matters
particularly for abstract names, where larger initial errors allow
larger absolute corrections. The four lexical regimes are descriptive
analysis groups, not explanatory semantic mechanisms.

\paragraph{The informative failure is miscalibrated action.}
Oracle corrections predictably help by revealing target information.
The stronger diagnostic is that directional compliance and ordered
modifier responses can coexist with harmful overshoots: these
measurements are incomplete proxies for target accuracy.
Receiver calibration partly addresses this mismatch without
establishing a model's internal mechanism or universal meanings
of modifiers. Its medians remain receiver- and prompt-specific,
and robustness against other fixed rules has not been tested.

\paragraph{Feedback must fit its receiver.}
The calibration results connect graded language to the update it elicits.
A directionally correct, ordered response can still overshoot; selecting
from measured steps addresses this mismatch without changing the phrase
inventory. The benefit holds against the tested fixed rule, not every
possible unfitted policy, and does not define universal modifier meanings
or human perceptual units. Independently fitted model/prompt mappings
support within-receiver calibration, not cross-model controller transfer.
The operational unit is the receiver's displayed action, rather than an
assumed numerical meaning of a word. Three phrase choices cannot express
arbitrary intermediate steps, and individual responses need not match
their calibration median. Numeric instructions add precision and coordinate
holds; their advantage does not isolate phrase resolution alone.

\paragraph{Evaluate messages through the updates they elicit.}
Teacher messages require both preservation and reception tests.
A faithful direction set leaves magnitude to the guesser, while additional
directions may help but violate the supplied information budget.
Comparative scope also matters: describing a guess as darker than a
target is different from requesting a darker guess. Joint audits of
messages and receiving updates distinguish these cases without treating
keyword overlap as a semantic judgment or task success as proof of fidelity.
These findings concern oracle-assisted pipelines, not an unaided
following-versus-generating capability gap.
Restricted zero-shot's favorable mean illustrates why both tests are
needed: it sometimes supplies more directions than the oracle arm, and
its gain is concentrated rather than uniform. Preserving the original
messages in reception exposes this difference; repairing them first would
change the interaction being evaluated.

\paragraph{State and stopping complete the control problem.}
Output encoding and its instructions shape correction. An axis legend
supports an instructed LAB interface, not unaided perceptual understanding.
Native and displayed states must remain distinct: sRGB projection can
alter movement and leave drift that a fixed-axis policy does not explicitly correct.
The three-phrase ladder also limits correction resolution. Even after a
close guess, further updates can lose convergence. Saved-prefix stopping
retains nine otherwise lost convergence outcomes and saves 23/576
shared-start revisions (Appendix~\ref{app:stopping}), but uses the hidden
target. Applying the game to human interaction or creative tools therefore
requires both richer revision control and a way to decide when a guess is
satisfactory without target access; neither is established here.
Description-based games test revision toward a recorded referent;
isolated tasks test wording-controlled adjustments. The sequential pilot
connects successive adjustments, but full calibrated, multi-axis games
remain untested.

\section{Conclusion and Future Work}
\label{sec:conclusion}
\textsc{ColorRef} studies color language as a reference game: a guesser
interprets a description, and a teacher with a recorded target guides
successive revisions through verbal feedback. The game makes initial
interpretation and response to correction separately observable. Its
controlled studies show that following a direction does not ensure a
useful step size.

Receiver-calibrated phrase selection addresses this mismatch by choosing
wording from separately measured adjustment behavior. Independently fitted
policies reduce held-out one-step mean error for Qwen3-14B and Gemma 4 12B
relative to the same fixed graded-wording rule, chiefly by avoiding
excessive large corrections. Many cases are unchanged, and threshold success
does not necessarily improve. Teacher controls further separate preservation
of supplied directions from receiving accuracy. These findings motivate
evaluating feedback through both its intended correction and the update
it elicits.

Next steps include calibrated feedback in full description-based games,
multi-axis revision, target-free stopping, and robustness to selection rules
and prompt wordings. Human and multilingual exchanges should account for
acceptable target variation.

\section{Limitations}
\label{sec:limitations}
Our recorded crowd targets need not be uniquely appropriate for their
descriptions. Both the retained corpus and raw English source record one
color per description, without repeated targets for estimating agreement
or acceptable variation. Neither corpus size nor validity filtering establishes
target agreement; alternative-color judgments or target distributions remain needed,
especially for abstract names. The four regimes are heuristic parser
assignments rather than independently validated semantic classes;
the balanced 1{,}000-item sample does not estimate source-corpus
frequencies. Reported between-regime gains are confounded by
initial error, without an initial-error-adjusted comparison.
CIELAB is approximately perceptually uniform, but Euclidean $\Delta E_{76}$ is not a human magnitude calibration or a
strict visibility threshold. Equal coordinate steps have equal norm in
this metric; differences between directions can also reflect generated
native steps and the restricted sRGB gamut, rather than perceptual
nonuniformity alone.

The magnitude confirmation covers two model families with the same prompt
and 128 held-out starting-color clusters. Other diagnostic controls use Qwen
alone and smaller fixed samples, including twelve-color pilot clusters.
The primary comparator is one fixed quantifier rule with cutpoints 9 and 18;
alternative unfitted rules, gradient-based selection policies, and
prompt paraphrases are not tested. We use $\Delta E_{76}$ as the
operational geometric error; CIEDE2000 sensitivity has not been
computed for these runs.
Only one frozen description sample is evaluated; independent resampling
would test sensitivity to sample selection.
The studies use English descriptions and supplied numeric or HEX-derived
starting states. Bootstrap intervals do not establish broad population
generalization, generation-randomness uncertainty, or multiplicity-adjusted
significance. Failed calibration and held-out outputs remain in counts;
matched parsed cohorts can introduce
selection bias. No outputs are repaired or filtered by projection cost.

Teacher studies test oracle-assisted restatement, not unaided geometry;
supplementary finite-grammar and assistant review are not independent human
labels. Restriction changes several prompt instructions simultaneously.
Numeric feedback adds exact magnitude and coordinate holds, and calibrated
wording adds a selected magnitude category; neither is information-matched
to bare directions. Target-known stopping and benchmark oracles are not
deployable model-only policies. Evaluation-call savings omit calibration
cost, and fixed-axis transfer does not intentionally correct off-axis drift.
Creative applications, multilingual behavior, HSV magnitude calibration,
and alternative perceptual metrics require further studies.

\begingroup
\let\aclthebibliography\thebibliography
\renewcommand{\thebibliography}[1]{%
  \aclthebibliography{#1}\raggedright}
\bibliography{references}
\endgroup
\appendix
\section{Corpus Construction and Examples}
\label{sec:appendix-dataset}
\label{app:corpus-examples}

The ColorNames-derived source pairs user-provided color names with sRGB HEX
codes. Preprocessing retains English strings with valid HEX codes and a
description-validity score $\geq 0.75$. The original validity scoring
uses GPT-4o to judge whether a string plausibly functions as a color description,
accepting literal, figurative, and subjective expressions while rejecting
spam, URLs, nonsensical strings, and personal messages. This score is not
a measure of inter-speaker agreement about the paired color. Canonicalization
assigns stable identifiers and normalizes fields; color conversion produces
RGB, HSV, and CIELAB coordinates. The final corpus has 518{,}830 rows.

A read-only inventory confirms that 43{,}187 retained rows have score exactly
0.75; a strict $>0.75$ filter would retain 475{,}643 rows instead.
Each of the 705{,}869 raw English entries and 518{,}830 retained entries has
a distinct description, including after lowercasing and whitespace
normalization. Neither inventory contains repeated description--color pairs
or multiple recorded colors for a description. The 725{,}203 evaluation
records likewise have distinct exact descriptions; their fields concern
language, reasoning, and quality score, without explicit votes or alternative
color targets. These inventories establish recorded multiplicity, not human
agreement: evaluation therefore measures distance from the assigned recorded
target, rather than from a uniquely correct interpretation.

\begin{table}[H]
\centering\small
\begin{tabular}{@{}lrr@{}}
\toprule
Regime & Count & Percent \\
\midrule
Abstract idiosyncratic & 238{,}416 & 46.0 \\
Prototype mediated & 137{,}420 & 26.5 \\
Compound associative & 94{,}246 & 18.2 \\
Explicit grounded & 48{,}748 & 9.4 \\
\bottomrule
\end{tabular}
\caption{Original corpus regime counts; rounded percentages need not sum
to exactly 100. Labels are coarse analysis categories.}
\end{table}

\begin{table}[H]
\centering\small
\begin{tabular}{@{}lp{0.41\columnwidth}l@{}}
\toprule
Regime & Description & Recorded target \\
\midrule
Explicit & faded old rose & \texttt{\#D2917A} \\
Prototype & montmorency cherry & \texttt{\#E13D2B} \\
Compound & cryogenic electric blue & \texttt{\#1926FF} \\
Abstract & woman of the dunes & \texttt{\#66615E} \\
\bottomrule
\end{tabular}
\caption{Illustrative descriptions and their recorded targets, retained
from the original manuscript. A target is not asserted to be the only
semantically plausible color.}
\end{table}

\paragraph{Structural parser.}
The deterministic parser lowercases strings, maps punctuation to spaces,
removes nonalphabetic characters, and collapses whitespace. It scans tokens
right-to-left for a basic-color head; preceding tokens are partitioned into
known modifiers and unknown prefix content. Confidence decreases with
unexplained prefixes. Parser v2 additionally normalizes selected \emph{-ish}
forms and plurals, allows prototype heads when no basic head is found, and
records prototype hits throughout the name. Hand-curated YAML lexicons and
explicitness, prototype, rarity, and abstraction scores determine the regime.
The parser is useful for reproducible stratification but has not been
validated as a human taxonomy or as a measure of semantic distance from
a literal color term. Categories are discrete summaries of lexical
patterns, and a description can plausibly mix literal and associative
content. The study neither tests the optimal number of categories nor
attributes model errors causally to membership in one of them. In
particular, the larger unadjusted gains for abstract descriptions may
reflect their larger initial recorded-target errors.

\section{Oracle Wording and Receiving Prompts}
\label{app:bandwidth_wording_examples}

The subscript $c1$, $c2$, or $c3$ specifies a maximum constraint count,
not a magnitude. For target \texttt{\#A0B0C0} and guess \texttt{\#805000},
the implementation selects muted, lighter, and blue in that order. These
examples are generated by the available code with example ID 0, seed 13,
and feedback turn 0; they are not recovered historical experiment messages.

\begin{center}
\small
\begin{tabular}{@{}lp{0.65\columnwidth}@{}}
\toprule
Condition & Example feedback \\
\midrule
Minimal $c1$ & Make it more muted. \\
Axis $c1$ & Make it more muted. \\
Axis $c2$ & Make it more muted and lighter. \\
Axis $c3$ & Make it more muted, lighter, and more blue. \\
Template $c1$ & Make it less saturated. \\
Template $c2$ & Make it less saturated. Move toward a brighter shade. \\
Template $c3$ & Make it less saturated. Move toward a brighter shade.
Move it toward blue. \\
\bottomrule
\end{tabular}
\end{center}

Template conditions use deterministic alternative phrases for the same
selected directions; the phrase bank does not add \emph{a little} or
\emph{much}. These modifiers belong to the separate magnitude studies.
Minimal and axis-$c1$ are equivalent in this controlled setting.
The template-vocab implementation still selects primitive LAB/HSV directions;
broader warmer/cooler or gray dictionary entries do not establish a matched
broad-vocabulary ceiling. No-candidate fallback messages can differ.

Complete HEX/LAB, controlled-adjustment, and teacher templates are
reproduced in Appendix~\ref{app:all_prompts}, together with the exact
four-shot demonstration inputs and variable-filling rules. The reported
game and receiver runs include a sentence identifying the previous guess
as the displayed sRGB state.

\section{State, Calibration, and Metric Details}
\label{app:framework_details}

\subsection{State protocols and trajectory interpretation}
\label{app:state_protocol_details}
The one-shot prediction is the initial prediction before revision; the
term does not specify the number of demonstrations in the prompt.
All generations use fresh contexts. In description-based games, a revision
contains the description, previous displayed guess, and latest feedback,
without the full conversation history.
The fresh grounding evaluation and shared-start interface control use
three fixed revisions: an initial or supplied state followed by turns 1--3.
They continue even when an intermediate guess is already close.
The sequential magnitude pilot instead stops at the first displayed
$\Delta E_{76}\leq5$, using the known target, and caps generation at five
revisions. This is neither a learned guesser stopping policy nor a universal
perceptual threshold.

A direct-LAB response retains native coordinates $z_t$ separately from
displayed $\hat c_t=P(z_t)$. Display conversion clips and rounds to uint8
sRGB, then converts back to LAB; a LAB revision prompt expresses this
displayed state in LAB coordinates. Rectangular LAB encoding bounds are
not the sRGB gamut, so native and displayed movement can differ.

In Figure~\ref{fig:colorref_overview}b, the first correction for
\textit{southern lime green} requests greater saturation, lower lightness,
and a shift toward green. Each new displayed guess becomes the input for
the next oracle comparison. Across three fixed revisions, full LAB error
falls from 32.8 to 8.8, but remains above five.
The example is selected for explanation, not as an estimate of typical
behavior; other saved trajectories regress or overshoot.
The lightness strip and full three-dimensional error prevent apparent
proximity in the $a^*b^*$ plane from being mistaken for target agreement.
Background colors are clipped sRGB at the target's fixed $L^*$, not a
gamut boundary; point fills and bars use the saved displayed colors.
Saturation feedback uses HSV, not a separate axis in that plane.
The oracle supplies at most three corrections per turn.

\subsection{Calibration selection details}
\label{app:calibration_selection_details}
The isolated calibration instructions request one of six LAB directions:
lighter, darker, more red, more green, more yellow, or more blue.
They separate adjustment behavior from predicting a color name and do not
test all multi-axis description-game dynamics.
Direction/phrase medians use parsed displayed signed steps, retaining
zero and wrong-direction updates. Failed parsing remains in completion
counts. Bare wording is a separate baseline and has no ordinal rank
or place among selection candidates.
The teacher selects the positive graded median closest to the requested-axis
target residual, breaking ties by smaller median and then fixed ladder
order. If no graded phrase is usable, the mapping is reported as unavailable.
Held-out responses never refit the mapping, and the receiving model's
weights are unchanged. Calibration medians are receiver- and prompt-specific.

\subsection{Diagnostic metric definitions}
\label{app:metric_definitions}
We use $\Delta E_{76}$, Euclidean distance in CIELAB, as the common
operational target-error scale. Unlike raw Euclidean RGB distance, this
scale separates a lightness coordinate from two opponent-color
coordinates and is designed to better approximate perceptual
differences. It remains an approximation: equal numerical distances
are not equally visible for every color pair, and sRGB clipping can
change both direction and magnitude. The metric is a reproducible
geometric endpoint for these experiments, not a validation of a
recorded color name or a threshold for human agreement.
CIEDE2000 could provide a complementary perceptual sensitivity
analysis, but is not reported here.

Besides initial and endpoint error and gain, mean per-example relative
improvement is $(e_0-e_T)/(e_0+\epsilon)$. It is not the percentage reduction
of the cohort mean.
For displayed update $u_t=\hat c_{t+1}-\hat c_t$ and ideal correction
$v_t=c-\hat c_t$, directional alignment is
\begin{equation}
A_t=\frac{v_t^\top u_t}{\|v_t\|_2\|u_t\|_2}.
\end{equation}
Zero-length vectors are excluded from alignment, not from target error.
Constraint satisfaction checks signed movement along supplied oracle
directions, including HSV saturation where applicable.

For an isolated instruction with unit requested direction $d$, native
signed step is $s=d^\top(z'-z)$ and off-axis drift is
$\|(z'-z)-sd\|_2$; displayed counterparts replace native states by their
displayed values. Positive steps follow the request, while zero and
wrong-sign steps are reported separately. Ordinal comparisons test
\emph{a little} $<$ \emph{somewhat} $<$ \emph{much}, with bare wording
unranked. Projection cost is $\|z'-P(z')\|_2$.

For a nonzero displayed update $u$ and target correction $v$,
\begin{equation}
\|v-u\|_2^2-\|v\|_2^2=\|u\|_2^2-2v^\top u.
\end{equation}
Thus a nonzero update reduces Euclidean error only when
$\|u\|_2<2\|v\|_2\cos\theta$, where $\theta$ is their angle.
A directionally aligned step can increase error by moving too far;
ordered graded responses can also leave a gap between available steps
and the correction needed. Sign compliance, magnitude ordering, and
target accuracy are distinct outcomes.

\section{Matched Controls and Reproducibility}
\label{app:new-controls}

\paragraph{How to read the comparisons.}
The description-based one-shot versus oracle-revision comparison asks
whether extra target-aware corrections help; the information budget
is not matched. The HEX-versus-LAB comparison holds starting displayed
colors and first correction fixed, but later feedback adapts to each
arm's revised color. The calibrated-versus-unfitted comparison holds
the target, starting color, receiver, instruction template, and
three-phrase vocabulary fixed while changing the phrase-selection
rule; it uses one revision and is the primary policy test. Bare
directions omit magnitude. Numeric feedback gives exact coordinate
changes and holds the remaining axes, so it tests execution rather
than a fair competition among equally informative phrasings.
Teacher restatement controls begin with supplied oracle directions;
they do not test unaided target inference by a language-model teacher.

\subsection{Grounding sample and oracle selection}
\label{app:grounding_protocol}
The fresh grounding sample contains 1{,}000 descriptions, 250 per regime,
drawn with seed 113 after excluding all 400 original debug-subset IDs.
Each game uses HEX output, an initial prediction and three fixed revisions,
with at most three canonical axis corrections per turn. The initial
prediction is also the paired one-shot baseline. Equal regime weights,
selected IDs, source/subset hashes, plans and prompts are frozen before
inference.

The deterministic oracle compares target and displayed guess along LAB
lightness, red--green, yellow--blue, and HSV saturation. Absolute
discrepancies must reach 2, 3, 3, and 0.05, respectively. Qualifying
candidates are ranked by absolute discrepancy divided by fixed ranges
100, 200, 220, and 1, and selected up to the configured bandwidth.
These ranges define a selection policy, not perceptual magnitude units.
Constraints are verbal property-direction requests, not hard restrictions
on the next output. Canonical wording contains no graded modifiers;
Appendix~\ref{app:bandwidth_wording_examples} distinguishes it from archived
template variants.

\subsection{Model execution and checkpoint settings}
\label{app:execution_settings}
Grounding and diagnostic controls use \texttt{Qwen/Qwen3-14B}, pinned to
revision \texttt{40c069824f4251a91eef}\allowbreak\texttt{af281ebe4c544efd3e18}.
The magnitude replication uses \texttt{google/gemma-4-12B-it}, with its
snapshot recorded below. Both run through Hugging Face Transformers in
bfloat16 on an A100 40\,GB, at temperature zero with thinking disabled.
Gemma uses its multimodal processor/model loader with text-only inputs.
Receiver output budgets are 64 tokens; teacher budgets are 80.
Plans, prompts, model-revision metadata and atomic response checkpoints
are preserved. Strictly invalid outputs are not repaired; backend
exceptions remain pending.

\subsection{Shared-start interface control}
The 64 descriptions are balanced across regimes. Starts come from the
completed parent HEX predictions, not new predictions by each receiving
interface. Target LAB is recomputed from recorded target HEX. The same
oracle thresholds and ranking operate on displayed uint8 sRGB states in
every arm. First corrections are identical; later corrections respond to
the arm's own displayed state. Three revisions are forced even after a
close guess, which differs from the original early-stopping protocol.
All 576 generated revisions parse, so first/final matched cohorts contain
all 64 examples. First alignment denominators are 64/61/63 for HEX/plain
LAB/legend LAB because some updates have zero length.

Final native errors are 22.781, 35.589, and 23.325; displayed errors are
22.781, 35.275, and 23.161. Mean projection errors over generated states
are 0, 0.783, and 0.937. Thus, an axis legend does not uniformly reduce
projection cost. These results must not be mixed with the earlier
own-initial-prediction interface study, whose common cohort is 61/64 after
three legend-condition parse failures.

\subsection{Isolated wording probes}
For each of 12 fixed colors and six requested directions, compare bare,
\emph{a little}, \emph{somewhat}, and \emph{much} under plain and legend
prompts. Each response is independent; the ordinal comparison does not
chain the resulting states. Exact 5-/10-unit controls explicitly name a
coordinate and hold the other two fixed. Twelve no-change cases per prompt
complete the 888-response plan. All responses parse. The qualitative ladder
has three pairwise comparisons per starting-color/direction combination,
216 comparisons per prompt. Bare wording has no ordinal rank.

\begin{table}[H]
\centering\small
\begin{tabular}{@{}lrr@{}}
\toprule
Measure & Plain & Legend \\
\midrule
Positive qualitative updates / 288 & 228 & 288 \\
Nondecreasing pairs / 216 & 208 & 216 \\
Strictly increasing pairs / 216 & 172 & 210 \\
Tied pairs / 216 & 36 & 6 \\
\bottomrule
\end{tabular}
\caption{Direction and ordinal magnitude in isolated probes.
Nondecreasing includes ties and does not imply correct-sign movement.}
\end{table}

There is no hidden target for qualitative probes. A numeric headroom fraction
uses the prompt's rectangular bounds and cannot measure a fraction of remaining
target distance or gamut capacity. Projection diagnostics are not response
filters. The one nonexact ordinal-study numeric control returns a 20-unit
yellow-axis update for an instruction specifying ten.

\subsection{Calibration policy and target construction}
Calibration and one-step evaluation use 12 colors each drawn from an inclusive
sRGB channel range [48,208], with seed 29. A subsequent seed-47 transfer split
contains twelve fresh starts, excluding the earlier calibration/evaluation
starts, isolated anchors, and the inspected close-case start by exact HEX.
Inputs and feasibility decisions are frozen before inference.

Calibration generates 12 colors $\times$ six directions $\times$ four
wordings, or 288 responses (287 parsed). For each direction and phrase,
take the median displayed signed step among parsed responses, retaining
zero/wrong-sign updates. Bare wording is not a selection candidate;
graded phrases with positive medians are. Select the median closest to the
current axis residual, breaking ties by smaller median and then ladder order.
No held-out response refits this mapping.

\begin{table}[H]
\centering\small
\begin{tabular}{@{}lrrr@{}}
\toprule
Direction & A little & Somewhat & Much \\
\midrule
Lighter & 2.949 & 7.961 & 28.722 \\
Darker & 4.001 & 10.013 & 30.118 \\
More red & 1.139 & 5.073 & 75.844 \\
More green & 1.277 & 6.620 & 30.038 \\
More yellow & 4.529 & 7.381 & 54.333 \\
More blue & 4.917 & 10.094 & 48.137 \\
\bottomrule
\end{tabular}
\caption{Frozen calibration medians in displayed requested-axis units.
These are operational, direction-specific measurements for this model/prompt,
not human or universal perceptual magnitudes.}
\end{table}

Native target candidates move one coordinate by 6, 12, or 24 units.
Retain candidates within the prompt bounds whose projection error is at
most one and whose displayed residual follows the intended direction.
This produces 196/216 one-step targets and 200/216 transfer targets;
excluded inputs are not replaced. Generated outputs remain unfiltered.
Displayed target construction and six-decimal prompted coordinates are
stored separately from exact instructed targets.

All arms use the same supplied start and target. Bare feedback specifies
direction. Calibrated feedback supplies a selected magnitude phrase.
Numeric feedback specifies an exact increment and holds the other coordinates;
it is a precision/execution control, not information-matched wording.
Sequential transfer retains the original axis throughout, stops at the
first displayed error at most five, and caps generation at five revisions.
If the axis residual is at most 0.01 while total error exceeds five,
the protocol permits an explicit off-axis-residual ending instead of an
empty correction. No such ending occurs in the completed transfer study.
Reaching the cap is a valid observed endpoint, not an imputed success.

\subsection{Four-arm magnitude confirmation}
The confirmation uses seed 211, 32 new calibration colors and 128 held-out
starting colors, excluding 49 previously used colors before inference.
Calibration covers six directions and four wordings (768 responses, 767
parsed). Frozen direction-specific medians use parsed responses including
nonpositive updates; held-out responses never refit the mapping.
The fixed unfitted rule selects \textit{a little} below residual 9,
\textit{somewhat} from 9 to below 18, and \textit{much} otherwise.
Calibrated wording instead selects the positive graded median nearest the
requested-axis residual. Bare directions and exact numeric instructions
complete the four arms. All use fresh contexts and one revision.

The existing native-distance targets at 6, 12 and 24 units yield 2{,}108
feasible cases; 196 candidates fail frozen input-only feasibility checks.
All 8{,}432 held-out responses complete. Parse failures are 3/5/0/2 for
bare/unfitted/calibrated/numeric. Primary means use 2{,}098 common parsed
quartets covering all 128 starts. The primary calibrated-minus-unfitted
contrast is $-5.497$ [$-5.786$, $-5.203$]; available paired responses
give $-5.528$ [$-5.814$, $-5.245$] on 2{,}103 cases.
All 12-unit and darker comparisons share identical prompts. On the common
cohort, 1{,}385 prompt-identical cases produce identical colors; the other
713 cases contain 556 wins, 34 ties and 123 losses at tolerance 0.01.
This identity partition is descriptive, not a replacement primary cohort.
Numeric execution has 39 parsed native errors above one, maximum 51.004;
misses remain in the reported performance.

\subsection{Gemma replication and saved-output diagnostics}
Gemma 4 12B uses revision \texttt{707f0a3b8a3c7ad586ed}\allowbreak\texttt{01e27eafbad8a27dd0f7},
the identical frozen plan and text prompts, and independently fitted medians.
All 768 calibration outputs parse. Of 8{,}432 evaluation responses, only
27 numeric outputs fail parsing; bare, unfitted and calibrated each parse
2{,}108/2{,}108. Common quartets contain 2{,}081 cases from all 128 starts.
The primary contrast is $-5.161$ [$-5.484$, $-4.842$]; available language
pairs give $-5.095$ [$-5.417$, $-4.772$] over all 2{,}108 cases.

Direction-specific medians differ between receivers under this protocol:
for \textit{a little more red}, displayed medians are 1.054 for Qwen
and 4.953 for Gemma. The experiment does not test a controller fitted on
one model and deployed on the other, so it establishes within-model
calibration benefits, not cross-model policy transfer.

The Gemma identity audit finds 1{,}547 identical prompts among planned
unfitted/calibrated pairs, with no parsed-LAB mismatch on identical prompts.
On common quartets, 1{,}524 identical prompts give zero error difference;
557 changed prompts give a mean difference of $-19.282$.
The overall median paired difference remains zero. Distance-specific effects
at 6/12/24 units are $-0.064$, $0.000$, and $-18.182$; the largest gains
occur for 24-unit red, yellow and blue targets. Harmful updates fall from
333 to 75, while threshold success remains nearly unchanged (1{,}400 to
1{,}398). All 128 per-start mean differences favor calibration; omitting
one start yields differences between $-5.194$ and $-5.121$.
These audits are exploratory and do not replace the primary cohort.

Among parsed numeric outputs, mean native execution error is 0.006;
2{,}058/2{,}081 are within 0.01 of instructed coordinates.
The 27 invalid outputs remain failures, without imputed colors or repaired
responses. Saved-output inspection finds all 27 failures in numeric
\textit{more green} instructions (decreasing $a$), across 17 starts:
four at distance 6, seven at 12 and sixteen at 24. Seventeen outputs
contain only two coordinates; ten contain a triplet rejected for negative
$L$. Twenty-five terminate at EOS below the budget, and two reach the
64-token cap after first emitting an invalid triplet and starting a correction.
Thus, truncation does not explain most observed failures.

For example, starting from $(49.742386,10.289397,5.133036)$ and decreasing
$a$ by 24.238567 requires $(49.742386,-13.949170,5.133036)$;
the model instead returns \texttt{LAB(-13.94917, 5.133036)}.
This example computes the requested $a$ value but omits $L$.
Other outputs place a negative adjusted value in the $L$ position or
duplicate coordinates. These are observed coordinate-output failures,
not evidence of a particular internal reasoning mechanism. No missing
coordinate is supplied retrospectively. Numeric parsing succeeds on
2{,}081/2{,}108 planned responses (98.7\%); its low parsed-response error
does not establish reliable execution across all planned cases.

\subsection{Confirmation contrasts and generation costs}
\label{app:confirmation_display}
\begin{figure*}[t]
\centering
\includegraphics[width=0.96\textwidth]{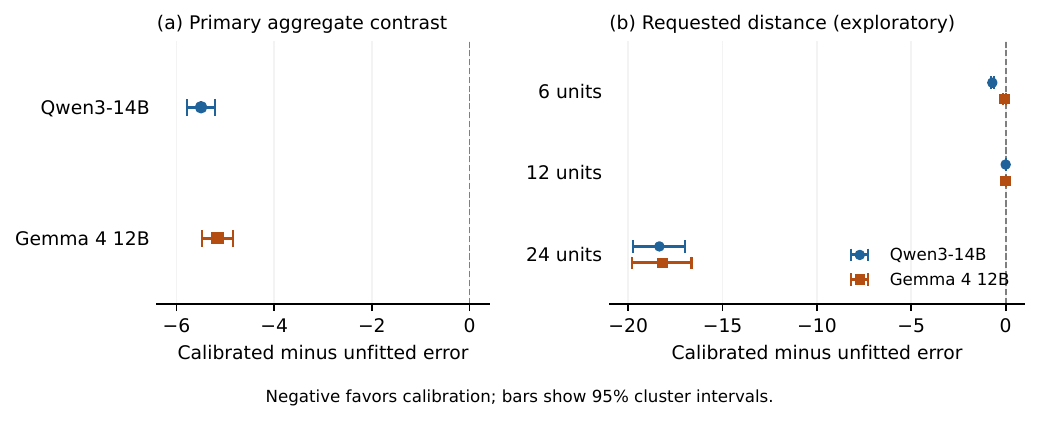}
\caption{Receiver-specific calibration reduces displayed target error relative
to the fixed unfitted quantifier rule for both models (left).
Exploratory distance breakdowns concentrate the benefit at 24-unit targets
(right). Bars show 95\% starting-color cluster percentile intervals;
each model covers the same 128 starting colors, with 2{,}098 Qwen and
2{,}081 Gemma common parsed quartets. These are within-model contrasts,
not a paired between-model test. The panels use different horizontal scales.
Values are reproduced from reported rounded estimates.}
\label{fig:calibration_effects}
\end{figure*}

The primary paired effects and exploratory distance breakdowns are visualized
in Figure~\ref{fig:calibration_effects}. Table~\ref{tab:magnitude_details}
retains completion-cohort means, available convergence counts, and
evaluation calls for the confirmation and separate sequential pilot.

\paragraph{Distance and policy-identity detail.}
For Qwen, calibrated-minus-unfitted error at requested distances 6, 12,
and 24 is $-0.71$, $0.00$, and $-18.35$, respectively; the 24-unit
calibrated mean remains 15.31. Harmful updates fall from 318 to 58 on
the common cohort. At tolerance 0.01, that cohort has 556 wins, 1{,}419
ties, and 123 losses, with median paired difference zero. Its 1{,}385
prompt-identical cases produce identical colors; the other 713 include
556 wins, 34 ties, and 123 losses. Calibration selects \textit{somewhat}
instead of \textit{much} for 24-unit red, yellow, and blue targets,
and instead of \textit{a little} for 6-unit lighter, red, and green
targets. All 12-unit and darker comparisons reuse identical prompts.

For Gemma, the 1{,}524 prompt-identical common cases produce identical
colors; the other 557 contain 443 wins, four ties, and 110 losses.
Calibration replaces \textit{much} with \textit{somewhat} for 24-unit
darker, red, yellow, and blue corrections, and usually replaces
\textit{a little} with \textit{somewhat} for 6-unit green corrections.
The distance effects, harmful-update counts, threshold outcomes, and
leave-one-start-out sensitivity are reported above. These post hoc
partitions localize the observed policy differences; they are not
randomized subgroups, independent controller experiments, or causal
effects of phrase choice. Subgroup intervals are exploratory.

\begin{table}[t]
\centering\small
\begin{tabular}{@{}llrrr@{}}
\toprule
Study & Arm & Final $\Delta E$ & $\leq5$ & Calls \\
\midrule
Qwen & Bare & 11.91 & -- & 2{,}108 \\
 & Unfitted & 12.36 & -- & 2{,}108 \\
 & Calibrated & 6.86 & -- & 2{,}108 \\
 & Numeric & 0.35 & -- & 2{,}108 \\
\midrule
Gemma & Bare & 8.36 & 924 & 2{,}108 \\
 & Unfitted & 11.06 & 1{,}400 & 2{,}108 \\
 & Calibrated & 5.89 & 1{,}398 & 2{,}108 \\
 & Numeric & 0.08 & 2{,}081 & 2{,}108 \\
\midrule
Sequential & Bare & 6.46 & 148 & 516 \\
 & Calibrated & 2.98 & 192 & 338 \\
 & Numeric & 0.10 & 200 & 205 \\
\bottomrule
\end{tabular}
\caption{One-step means on model-specific common quartets: 2{,}098 for
Qwen and 2{,}081 for Gemma, each spanning the same 128 starts.
Convergence counts use each arm's parsed responses, not the common quartets.
Calls include failed outputs and exclude 768 calibration calls per model.
Sequential Qwen pilot: 200 targets from 12 further starts, stopping at
error $\leq5$ with a five-revision cap, excluding 288 calibration calls.
Numeric feedback supplies additional precision.}
\label{tab:magnitude_details}
\end{table}

\subsection{Uncertainty and saved inputs}
Paired bootstrap draws resample whole starting colors for magnitude studies
and whole examples within regimes for interface/teacher studies, keeping
associated direction/distance or bandwidth cases together. Analysis settings
are retained with the saved run metrics. Pooled case weights remain unchanged. The pilot
uses twelve starting-color clusters; the confirmation uses 128. Multiple
targets per start are not independent colors. Available-pair sensitivity
equals common-triplet estimates in the fully parsed pilot; confirmation
quartets and available pairs have different, explicitly reported denominators.

Frozen configuration, plans, calibration/controller snapshots, rendered
prompts, responses, parsing results, native/displayed states and scores are
retained. Resume validates hashes, scores and observed model revision; backend
failures stay pending, while invalid generated outputs remain failures.
The observed model snapshot is recorded in run metadata. CPU inspectors
write supplementary reports without changing checkpoints or fitting policies.
Software tests use simulated outputs and do not supply model-performance evidence.

\section{Magnitude and Stopping Diagnostics}
\label{app:magnitude_diagnostics}

\subsection{One-step limitations}
Calibrated versus bare has 93 wins, 57 ties and 46 losses on 196 targets
at an exploratory 0.01 $\Delta E$ tie tolerance. Mean differences for
6-/12-/24-unit targets are $-9.84$, $-5.77$, and $+0.85$, respectively;
the 24-unit interval spans zero. The red calibration ladder has a large gap
between \emph{somewhat} (5.073) and \emph{much} (75.844) displayed units.
This diagnoses coarse resolution, not an identified internal linguistic
mechanism. Per-start leave-one-out mean effects remain favorable, but are
descriptive leverage checks rather than new intervals or grounds for exclusion.

The exact numeric control has 188/196 native responses within 0.01 of the
instructed coordinates and 192 within one. Four larger misses are 39, 32,
12, and 10 units. They terminate at EOS below the output budget and preserve
unrequested coordinates, with projection displacement below 0.5. They remain
in reported performance. For example, an instructed red-axis change from
$-43.698223$ to $-37.939764$ instead returns $1.060236$, a 39-unit miss.
Observed execution/transcription failure does not prove a particular internal
digit or sign-loss mechanism.

\subsection{Sequential prefixes and failures}
All rows below concern the same 200 transfer targets. Earlier stopped states
are retained with no additional calls. Prefixes are saved observations,
not independently rerun or optimized stopping policies.

\begin{table}[H]
\centering\small
\begin{tabular}{@{}lrrr@{}}
\toprule
Budget & Bare & Calibrated & Numeric \\
\midrule
1 & 10.147 & 6.434 & 0.534 \\
3 & 7.006 & 3.339 & 0.095 \\
5 & 6.461 & 2.979 & 0.095 \\
\bottomrule
\end{tabular}
\caption{Mean displayed target error at observed transfer prefixes under
the frozen five-revision protocol.}
\end{table}

Calibrated convergence counts are 119, 187, and 192 at these budgets.
The fourth/fifth revisions add five successes at 22 additional calls;
the primary endpoint remains the declared five-revision cap.
Bare/calibrated trajectories contain 152/11 harmful revisions and 174/48
requested-sign reversals. Their unequal adaptive histories make these
descriptive counts, not independent per-step significance samples.

All eight calibrated cap endings concern 24-unit lighter/greener corrections
on six starting colors. Their final off-axis disagreement exceeds five;
seven also have requested-axis residual below five. Large native proposals
project into shifted unrequested coordinates. The fixed-axis controller
does not explicitly repair that drift; it is not universally impossible
for later projection to change unrequested coordinates incidentally.
Bare converges on six of these eight targets and numeric on all eight.

Numeric transfer has 199/205 native revisions within 0.01 and 200 within one,
with maximum native error 44. Five large first-revision errors reach the
threshold after a second target-known correction; one further small error
is below the stopping threshold. Endpoint success cannot erase intermediate
execution failures or establish autonomous self-correction.

\subsection{Target-known stopping replay}
\label{app:stopping}
Replay the shared-start trajectories, retaining the first displayed state
with $\Delta E_{76}\leq5$, including assigned starts. This performs no
new generation and does not select the retrospective best state.

\begin{table}[H]
\centering\small
\begin{tabular}{@{}lrrrr@{}}
\toprule
Interface & $e_T$ & $e_{\rm stop}$ & $n_T$ & $n_{\rm stop}$ \\
\midrule
HEX & 22.781 & 22.005 & 192 & 185 \\
Plain LAB & 35.275 & 34.562 & 192 & 186 \\
LAB + legend & 23.161 & 22.822 & 192 & 182 \\
\bottomrule
\end{tabular}
\caption{Stopping replay on the same 64 examples in each interface.
Columns $e$ and $n$ give mean target error and generation calls,
respectively, for fixed and stopped policies.}
\end{table}

Nine otherwise lost convergence outcomes occur on seven distinct examples;
three share the initially close \emph{dull toy red} start.
Its starting error is 2.380, yet valid-axis first corrections produce gains
of $-27.646$, $-47.017$, and $-33.298$ in HEX/plain LAB/legend LAB.
These are normally terminated outputs with low projection cost: magnitude,
not simply geometric sign, can destroy reached success. This single case
is not a near-target failure-rate estimate. Stopping is benchmark target-known,
not a learned or deployable model-only policy.

\section{Teacher Prompts, Audits, and Reception}
\label{app:teacher_controls}

The natural zero-shot condition preserves the archived oracle-assisted
prompt: it supplies correct directions as JSON and asks for a natural short
comparative paraphrase. Its vocabulary includes LAB/HSV primitives as well
as warmer/cooler and gray terms. The restricted prompt instead requests
exactly the supplied primitive directions, prohibits additions, omissions,
reversals, substitutions and magnitude modifiers, and requires a single
\emph{Make it} sentence. This changes multiple instructions and the allowed
format simultaneously; it does not isolate one individual wording instruction.

Four-shot prompts use canonical oracle answers for four input-selected
demonstrations. IDs are 202953, 207855, 39272, and 352171; evaluation excludes
these IDs and normalized description or target/guess-pair overlap. The same
IDs and order appear in both forms, with bandwidth-specific answers.
The final 60 examples contain 15 per regime; each is evaluated at $c1$ and
$c3$. The clause budget equals the maximum supplied direction count and
never forces lossy compression. Demonstrations do not cover every direction:
darker and more saturated are absent from their logged primitive coverage.

All teacher conditions use Qwen3-14B, temperature zero, thinking disabled,
and an 80-token output cap; the HEX receiver uses the same model family
and a 64-token cap. Four-shot prompts necessarily cost more input tokens.
The five receiving arms use every message unchanged, with no target-HEX
mentions observed and no parse failures in 600 revisions. There is no new
teacher generation for the deterministic oracle arm.

\subsection{Frozen lexical audit versus scope-aware diagnostics}
Primary conservative certification requires an exact detected direction set,
without unknown or ambiguous wording or duplicate mentions. It is an
operational format/vocabulary criterion, not semantic ground truth.
Supplementary finite-grammar auditing resolves explicit guess-versus-target
comparisons and a leading \emph{Feedback:} label while retaining unresolved
language for review. It does not overwrite primary scores.

\begin{table}[H]
\centering\small
\begin{tabular}{@{}lrrr@{}}
\toprule
Condition & Certified & Match & Review \\
\midrule
Natural, 0-shot & 7 & 81 & 22 \\
Natural, 4-shot & 3 & 120 & 0 \\
Restricted, 0-shot & 104 & 104 & 0 \\
Restricted, 4-shot & 105 & 120 & 0 \\
\bottomrule
\end{tabular}
\caption{Counts out of 120 teacher cases; Match denotes resolved matches
and Review denotes unresolved cases. Certification and resolved-match
scores are different measures; failures of certification are not necessarily
wrong corrections. Natural zero-shot also has 17 resolved mismatches.}
\end{table}

For example, the response \emph{Your guess is darker than the target color}
contains \emph{darker} but implies a lighter correction. In contrast,
\emph{Feedback: Make it lighter} directly preserves that direction while
violating strict canonical formatting. Contextual assistant review of the
22 unresolved cases is retained separately; it is neither independent human
annotation nor a replacement for the finite-grammar or primary metrics.
No review label selects teacher messages for receiving evaluation.

\subsection{Paired receiving effects and saved-message decomposition}
All receiver contrasts retain 120 cases on 60 whole-example clusters,
with both bandwidths together. Selected teacher-minus-oracle paired effects
are given below; intervals are exploratory and not equivalence tests.

\begin{table}[H]
\centering\small
\begin{tabular}{@{}lrr@{}}
\toprule
Arm & $\Delta$ error & 95\% interval \\
\midrule
Natural, 0-shot & 3.110 & [$-1.491$, 7.122] \\
Natural, 4-shot & 0.619 & [$-1.794$, 2.686] \\
Restricted, 0-shot & $-1.841$ & [$-3.772$, $-0.322$] \\
Restricted, 4-shot & $-0.104$ & [$-0.397$, 0.085] \\
\bottomrule
\end{tabular}
\caption{Displayed target-error differences; negative favors the teacher arm.}
\end{table}

Restricted zero-shot versus oracle partitions into 98 identical messages
(all ties), six different messages preserving the lexical direction set
(sum error difference $-126.040$; 3 wins, 1 tie, 2 losses), and sixteen
messages with additions (sum $-94.853$; 8 wins, 5 ties, 3 losses).
Together these yield the observed mean $-1.841$, but median zero.
The 57.1\%/42.9\% contributions describe net arithmetic effects, not a
randomized test of ordering, wording, or added information. The six same-set
messages comprise four reorderings (3 wins, 1 loss) and two punctuation or
conjunction changes (1 tie, 1 loss); all occur at c3. In example 52032,
\textit{Make it darker, more green, and more yellow} produces
\texttt{\#003366}, whereas the oracle's
\textit{Make it more yellow, more green, and darker} produces
\texttt{\#006600}. Teacher-minus-oracle error is $-92.008$, about 73\% of
the six cases' net advantage. These saved matched outputs demonstrate
sensitivity to equivalent correction forms, but do not identify an internal
processing mechanism or establish a generally better direction order.

Eight addition cases use only threshold-qualified extra primitive directions;
eight do not. Threshold qualification is neither sufficient for useful
magnitude nor necessary for a direction to have a correct sign. Some
contradictory primitive messages still improve target error, underscoring the
distinction between fidelity and usefulness. Natural-language keyword
additions are scope-sensitive and are not semantic bandwidth violations.
Across all arms, identical actual prompts never yield different parsed
colors in the saved audit; this is observed consistency, not a proof of
backend determinism. Restricted four-shot ties the oracle on 118/120 cases.


\section{Prompt Templates and Instantiation}
\label{app:all_prompts}
This appendix reproduces the task templates for the reported generation
studies and specifies their variable fields, adjustment instructions,
and four-shot teacher inputs. Display line wrapping is typographic;
source prompts retain their blank lines and final newline.
Each generation supplies one user-role message to the pinned model's
chat template, with generation-prompt insertion and thinking disabled.
No separate authored system message is added. Four-shot examples are
text inside that user message, not earlier conversation turns.
The model-specific chat formatting is supplied by its tokenizer or
processor at the recorded revision (Appendix~\ref{app:execution_settings}).

\begin{table}[H]
\centering\small
\begin{tabular}{@{}p{.43\columnwidth}p{.49\columnwidth}@{}}
\toprule
Study & Task templates \\
\midrule
Fresh grounding and its one-shot baseline & HEX initial and revision \\
Shared-start interfaces & HEX / plain LAB / axis-legend LAB revision \\
Earlier own-initial interface diagnostic & All three initial/revision pairs \\
Isolated wording probes & Plain and axis-legend LAB adjustment \\
Magnitude pilot, confirmation, Gemma replication, and sequential transfer & Axis-legend LAB adjustment \\
Teacher restatement & Natural or restricted teacher, zero-shot or four-shot \\
Feedback reception & HEX revision with original teacher text \\
Stopping replay and CPU audits & No new model generation \\
\bottomrule
\end{tabular}
\caption{Prompt coverage. Assigned shared starts are not initial generations.}
\end{table}

\subsection{Description-based guesses and revisions}
\label{app:game_prompts}
\texttt{\{raw\_name\}} is the corpus description without normalization
for display. The template adds quotation marks around it.
\texttt{\{previous\_guess\}} is the displayed state's HEX code for HEX
conditions, or \texttt{LAB(L, a, b)} with six decimal places for LAB
conditions. \texttt{\{feedback\}} contains the latest message only.
No target, constraint JSON, condition label, or earlier history is
appended to a guesser prompt. The initial template is also the
description-only baseline; shared-start controls use only revision
templates after assigning their common start.

\paragraph{HEX initial prediction}
\label{app:prompt_hex_initial}
\begin{quote}\footnotesize\ttfamily\raggedright
\detokenize{You are playing a color guessing game.}\par
\smallskip
\detokenize{Color description:}\par
\detokenize{"{raw_name}"}\par
\smallskip
\detokenize{Guess the intended color.}\par
\smallskip
\detokenize{Answer only with a valid hex code in the form #RRGGBB.}\par
\end{quote}

\paragraph{HEX revision}
\label{app:prompt_hex_revision}
\begin{quote}\footnotesize\ttfamily\raggedright
\detokenize{You are playing a color guessing game.}\par
\smallskip
\detokenize{Color description:}\par
\detokenize{"{raw_name}"}\par
\smallskip
\detokenize{Your previous guess was: {previous_guess}}\par
\detokenize{The previous guess is the displayable sRGB color used by the teacher.}\par
\detokenize{Feedback: {feedback}}\par
\smallskip
\detokenize{Revise your guess.}\par
\smallskip
\detokenize{Answer only with a valid hex code in the form #RRGGBB.}\par
\end{quote}

The HEX revision includes the sentence identifying the previous guess
as the displayable sRGB state. In feedback reception,
\texttt{\{feedback\}} is the teacher's complete original response, without
semantic repair or audit labels; the oracle arm uses its canonical message.

\paragraph{Plain LAB initial prediction}
\label{app:prompt_lab_initial}
\begin{quote}\footnotesize\ttfamily\raggedright
\detokenize{You are playing a color guessing game.}\par
\smallskip
\detokenize{Color description:}\par
\detokenize{"{raw_name}"}\par
\smallskip
\detokenize{Guess the intended color.}\par
\smallskip
\detokenize{Use CIELAB coordinates (D65): L in [0, 100], a and b in [-128, 127].}\par
\detokenize{Answer only in the form LAB(L, a, b), using numeric values.}\par
\end{quote}

\paragraph{Plain LAB revision}
\label{app:prompt_lab_revision}
\begin{quote}\footnotesize\ttfamily\raggedright
\detokenize{You are playing a color guessing game.}\par
\smallskip
\detokenize{Color description:}\par
\detokenize{"{raw_name}"}\par
\smallskip
\detokenize{Your previous guess was: {previous_guess}}\par
\detokenize{The previous guess is the displayable sRGB color used by the teacher.}\par
\detokenize{Feedback: {feedback}}\par
\smallskip
\detokenize{Revise your guess.}\par
\smallskip
\detokenize{Use CIELAB coordinates (D65): L in [0, 100], a and b in [-128, 127].}\par
\detokenize{Answer only in the form LAB(L, a, b), using numeric values.}\par
\end{quote}

\paragraph{Axis-legend LAB initial prediction}
\label{app:prompt_legend_initial}
\begin{quote}\footnotesize\ttfamily\raggedright
\detokenize{You are playing a color guessing game.}\par
\smallskip
\detokenize{Color description:}\par
\detokenize{"{raw_name}"}\par
\smallskip
\detokenize{Guess the intended color.}\par
\smallskip
\detokenize{CIELAB axis directions: increasing L makes the color lighter and decreasing L makes it darker; increasing a makes it more red and decreasing a makes it more green; increasing b makes it more yellow and decreasing b makes it more blue.}\par
\smallskip
\detokenize{Use CIELAB coordinates (D65): L in [0, 100], a and b in [-128, 127].}\par
\detokenize{Answer only in the form LAB(L, a, b), using numeric values.}\par
\end{quote}

\paragraph{Axis-legend LAB revision}
\label{app:prompt_legend_revision}
\begin{quote}\footnotesize\ttfamily\raggedright
\detokenize{You are playing a color guessing game.}\par
\smallskip
\detokenize{Color description:}\par
\detokenize{"{raw_name}"}\par
\smallskip
\detokenize{Your previous guess was: {previous_guess}}\par
\detokenize{The previous guess is the displayable sRGB color used by the teacher.}\par
\detokenize{Feedback: {feedback}}\par
\smallskip
\detokenize{Revise your guess.}\par
\smallskip
\detokenize{CIELAB axis directions: increasing L makes the color lighter and decreasing L makes it darker; increasing a makes it more red and decreasing a makes it more green; increasing b makes it more yellow and decreasing b makes it more blue.}\par
\smallskip
\detokenize{Use CIELAB coordinates (D65): L in [0, 100], a and b in [-128, 127].}\par
\detokenize{Answer only in the form LAB(L, a, b), using numeric values.}\par
\end{quote}

\subsection{Deterministic oracle messages}
\label{app:oracle_message_forms}
The oracle's selected direction phrases are \textit{lighter},
\textit{darker}, \textit{more red}, \textit{more green}, \textit{more yellow},
\textit{more blue}, \textit{more saturated}, and \textit{more muted}.
Preserve the ranking order selected by the policy in
Appendix~\ref{app:grounding_protocol}. The exact one-, two-, and three-direction
forms inserted into \texttt{\{feedback\}} are:

\begin{quote}\footnotesize\ttfamily\raggedright
\detokenize{Make it {direction_1}.}\par
\detokenize{Make it {direction_1} and {direction_2}.}\par
\detokenize{Make it {direction_1}, {direction_2}, and {direction_3}.}\par
\end{quote}

If no candidate reaches its selection threshold, the fixed-round
game inserts the following message:

\begin{quote}\footnotesize\ttfamily\raggedright
\detokenize{Your guess looks close. Try again.}\par
\end{quote}

These are deterministic text outputs, not teacher-model prompts.
The archived minimal/template illustrations in
Appendix~\ref{app:bandwidth_wording_examples} are separate wording variants.

\subsection{Controlled LAB adjustment templates}
\label{app:adjustment_prompts}
The isolated wording study fills \texttt{\{base\_lab\}} with its supplied
LAB anchor at two decimal places. Magnitude calibration, held-out
evaluation, and sequential transfer use displayed LAB at six decimal
places. All arms of the latter studies use the axis-legend template,
including bare and numeric controls. The same text templates are used
for Qwen and Gemma; each fits its own phrase-selection mapping.

\paragraph{Plain adjustment}
\label{app:prompt_adjustment_plain}
\begin{quote}\footnotesize\ttfamily\raggedright
\detokenize{You are applying a controlled color adjustment.}\par
\smallskip
\detokenize{Current color in CIELAB (D65): {base_lab}}\par
\detokenize{Instruction: {instruction}}\par
\smallskip
\detokenize{Return the resulting color after applying the instruction.}\par
\detokenize{Answer only in the form LAB(L, a, b), using numeric values.}\par
\detokenize{Keep L in [0, 100] and a and b in [-128, 127].}\par
\end{quote}

\paragraph{Axis-legend adjustment}
\label{app:prompt_adjustment_legend}
\begin{quote}\footnotesize\ttfamily\raggedright
\detokenize{You are applying a controlled color adjustment.}\par
\smallskip
\detokenize{CIELAB axis convention:}\par
\detokenize{- L: increasing makes the color lighter; decreasing makes it darker.}\par
\detokenize{- a: increasing shifts toward red; decreasing shifts toward green.}\par
\detokenize{- b: increasing shifts toward yellow; decreasing shifts toward blue.}\par
\smallskip
\detokenize{Current color in CIELAB (D65): {base_lab}}\par
\detokenize{Instruction: {instruction}}\par
\smallskip
\detokenize{Return the resulting color after applying the instruction.}\par
\detokenize{Answer only in the form LAB(L, a, b), using numeric values.}\par
\detokenize{Keep L in [0, 100] and a and b in [-128, 127].}\par
\end{quote}

\subsection{Adjustment instruction variants}
\label{app:adjustment_instructions}
The exact graded instruction form is:

\begin{quote}\footnotesize\ttfamily\raggedright
\detokenize{Make it {modifier}{direction}.}\par
\end{quote}

For bare wording, \texttt{\{modifier\}} is empty. Otherwise it is
exactly \texttt{a little }, \texttt{somewhat }, or \texttt{much }, including
the following space. \texttt{\{direction\}} is one of \texttt{lighter},
\texttt{darker}, \texttt{more red}, \texttt{more green}, \texttt{more yellow},
or \texttt{more blue}. Thus the rule specifies all 24 direction/wording
combinations, including \texttt{Make it a little more red.}
Bare, unfitted, and calibrated are experimental arm names, not text shown
to the receiver; the latter two differ only in the selected modifier.

The exact numeric form is:

\begin{quote}\footnotesize\ttfamily\raggedright
\detokenize{{verb} the {coordinate} coordinate by exactly {step} units. Leave the other coordinates unchanged.}\par
\end{quote}

\texttt{\{verb\}} is \texttt{Increase} or \texttt{Decrease};
\texttt{\{coordinate\}} is \texttt{L}, \texttt{a}, or \texttt{b}.
Lighter/red/yellow increase L/a/b; darker/green/blue decrease them.
In the isolated study, \texttt{\{step\}} is the integer text \texttt{5}
or \texttt{10}. Its no-change control is:

\begin{quote}\footnotesize\ttfamily\raggedright
\detokenize{Keep the current color exactly unchanged.}\par
\end{quote}

In magnitude evaluation, the numeric step has six decimal places.
It is the absolute difference between the displayed target and starting
coordinates, each rounded to six decimals, on the requested axis.
It need not equal the native target-construction distance 6, 12, or 24.
Sequential transfer recomputes this instruction from the current displayed
state and residual sign, retaining the original axis. All arms receive
only their instruction and current color, without the hidden target or
fitted medians. A threshold stop, exhausted budget, or unavailable mapping
does not generate a replacement prompt.

\subsection{Oracle-assisted teacher templates}
\label{app:teacher_prompts}
Unlike the guesser, the teacher receives \texttt{\{target\_hex\}} and
\texttt{\{guess\_hex\}}, together with \texttt{\{raw\_name\}}.
\texttt{\{max\_clauses\}} is the integer bandwidth 1 or 3.
\texttt{\{oracle\_constraints\_json\}} is a JSON array of selected
corrections, preserving their order. Each object has only
\texttt{direction} followed by \texttt{axis}, formatted with two-space
indentation; signed deltas and target coordinates are not additional fields.
For example, the first three-direction demonstration supplies:

\begin{quote}\footnotesize
\begin{verbatim}
[
  {
    "direction": "more muted",
    "axis": "hsv_s"
  },
  {
    "direction": "lighter",
    "axis": "lab_l"
  },
  {
    "direction": "more green",
    "axis": "lab_a"
  }
]
\end{verbatim}
\end{quote}

Direction-to-axis mappings are lightness to \texttt{lab\_l},
red/green to \texttt{lab\_a}, yellow/blue to \texttt{lab\_b}, and
saturated/muted to \texttt{hsv\_s}. Zero-shot conditions use the
following templates directly.

\paragraph{Natural paraphrase teacher}
\label{app:prompt_teacher_natural}
\begin{quote}\footnotesize\ttfamily\raggedright
\detokenize{You are the teacher in a color reference game.}\par
\smallskip
\detokenize{Target color description:}\par
\detokenize{"{raw_name}"}\par
\smallskip
\detokenize{True target color: {target_hex}}\par
\detokenize{Current guess: {guess_hex}}\par
\smallskip
\detokenize{The correct perceptual corrections for this step are (paraphrase only, do not list as JSON):}\par
\detokenize{{oracle_constraints_json}}\par
\smallskip
\detokenize{Turn these into one natural short sentence of comparative feedback for the guesser.}\par
\smallskip
\detokenize{Use only these feedback types:}\par
\detokenize{- lighter or darker}\par
\detokenize{- more red or more green}\par
\detokenize{- more yellow or more blue}\par
\detokenize{- more saturated or more muted}\par
\detokenize{- warmer or cooler}\par
\detokenize{- closer to gray or less gray}\par
\smallskip
\detokenize{Use at most {max_clauses} feedback clause(s).}\par
\smallskip
\detokenize{Do not reveal the target hex code.}\par
\detokenize{Do not mention exact RGB or LAB values.}\par
\detokenize{Do not mention a new object example.}\par
\detokenize{Do not say the exact answer.}\par
\detokenize{Answer with one short sentence only.}\par
\end{quote}

\paragraph{Restricted exact-direction teacher}
\label{app:prompt_teacher_restricted}
\begin{quote}\footnotesize\ttfamily\raggedright
\detokenize{You are the teacher in a color reference game.}\par
\smallskip
\detokenize{Target color description:}\par
\detokenize{"{raw_name}"}\par
\smallskip
\detokenize{True target color: {target_hex}}\par
\detokenize{Current guess: {guess_hex}}\par
\smallskip
\detokenize{The correct perceptual corrections for this step are:}\par
\detokenize{{oracle_constraints_json}}\par
\smallskip
\detokenize{State exactly the supplied directions. Do not add, omit, reverse, or substitute any direction.}\par
\detokenize{Use only these exact direction phrases:}\par
\detokenize{lighter, darker, more red, more green, more yellow, more blue, more saturated, more muted.}\par
\detokenize{Do not use warmer, cooler, gray terms, magnitude modifiers, or object examples.}\par
\detokenize{Use at most {max_clauses} directional clause(s).}\par
\detokenize{Return one sentence beginning "Make it", followed by every supplied direction exactly once, joined by commas and/or "and", and ending with a period.}\par
\detokenize{Do not reveal the target hex code or any RGB or LAB values.}\par
\detokenize{Answer with the feedback sentence only; do not list JSON or explain your reasoning.}\par
\end{quote}

\subsection{Four-shot teacher demonstrations and assembly}
\label{app:teacher_demonstrations}
Both teacher forms use the same four demonstrations, in the order below.
Their descriptions and HEX inputs are recovered from saved four-shot
prompts. The inputs are unchanged between bandwidths; their supplied
directions and canonical answers differ. Demonstration answers are oracle
sentences, not model-generated outputs.

\begin{table}[H]
\centering\small
\begin{tabular}{@{}p{.57\columnwidth}p{.35\columnwidth}@{}}
\toprule
Demonstration & Teacher inputs \\
\midrule
1. green leaf tea \newline ID 202953 & Target \texttt{\#6ae868} \newline Guess \texttt{\#008000} \\
2. grouchy cloud blue \newline ID 207855 & Target \texttt{\#5f7e72} \newline Guess \texttt{\#4a5568} \\
3. blinding green shade \newline ID 39272 & Target \texttt{\#67f479} \newline Guess \texttt{\#00ff00} \\
4. pool of acid \newline ID 352171 & Target \texttt{\#ff75e1} \newline Guess \texttt{\#00ff00} \\
\bottomrule
\end{tabular}
\caption{Complete description, target, and guess fields for the four-shot demonstrations.}
\end{table}

\begin{table}[H]
\centering\small
\begin{tabular}{@{}lp{.80\columnwidth}@{}}
\toprule
Demo & Canonical answer \\
\midrule
1 & Make it more muted. \\
2 & Make it lighter. \\
3 & Make it more muted. \\
4 & Make it more red. \\
\bottomrule
\end{tabular}
\caption{Bandwidth 1 demonstration answers. The selected JSON directions follow the same order as these sentences.}
\end{table}

\begin{table}[H]
\centering\small
\begin{tabular}{@{}lp{.80\columnwidth}@{}}
\toprule
Demo & Canonical answer \\
\midrule
1 & Make it more muted, lighter, and more green. \\
2 & Make it lighter, more green, and more yellow. \\
3 & Make it more muted, more blue, and more red. \\
4 & Make it more red, more blue, and more muted. \\
\bottomrule
\end{tabular}
\caption{Bandwidth 3 demonstration answers. The selected JSON directions follow the same order as these sentences.}
\end{table}

Each demonstration fills the selected teacher template with
description, target, guess, bandwidth, and JSON directions.
The exact wrapper below inserts rendered templates at
\texttt{\{demo\_n\}} and \texttt{\{evaluation\_case\}},
restoring blank lines and the final newline.

\begin{quote}\footnotesize\ttfamily\raggedright
\detokenize{Worked examples:}\par
\smallskip
\detokenize{Example 1:}\par
\detokenize{{demo_1}}\par
\smallskip
\detokenize{Feedback: {answer_1}}\par
\smallskip
\detokenize{Example 2:}\par
\detokenize{{demo_2}}\par
\smallskip
\detokenize{Feedback: {answer_2}}\par
\smallskip
\detokenize{Example 3:}\par
\detokenize{{demo_3}}\par
\smallskip
\detokenize{Feedback: {answer_3}}\par
\smallskip
\detokenize{Example 4:}\par
\detokenize{{demo_4}}\par
\smallskip
\detokenize{Feedback: {answer_4}}\par
\smallskip
\detokenize{New case:}\par
\detokenize{{evaluation_case}}\par
\end{quote}

Both forms repeat their full instructions within each demonstration
and evaluation case, without assistant-role examples.
Demonstrations and overlapping examples are excluded from evaluation
(Appendix~\ref{app:teacher_controls}).

\subsection{Upstream corpus-score prompt availability}
The builder uses precomputed description-validity scores, but its
exports do not include the original upstream scoring prompt.
We do not reconstruct that unavailable prompt from derived fields.
The templates above cover the reported generation experiments.

\section{AI Assistance Disclosure}
\label{app:ai_assistance}
Claude Code assisted experiment management and ablations; Codex assisted
refactoring, tests, analysis scripts, and manuscript drafting.
AI-assisted contextual review of teacher messages was not independent
human annotation and neither replaced frozen audit scores nor selected
evaluation cases. GPT-4o supplied the source description-validity scores.
The authors checked the experimental design, outputs, analyses, and manuscript.

\end{document}